\documentclass{article}

\usepackage{fancybox}
\usepackage{framed}
\usepackage{listings}

\usepackage{algorithm}
\usepackage{algorithmic}
\usepackage{subfigure}
\usepackage{subcaption}
\usepackage[dvipsnames]{xcolor}

\usepackage{PRIMEarxiv}

\usepackage[utf8]{inputenc} %
\usepackage[T1]{fontenc}    %
\usepackage{hyperref}       %
\usepackage{url}            %
\usepackage{booktabs}       %
\usepackage{amsfonts}       %
\usepackage{nicefrac}       %
\usepackage{microtype}      %
\usepackage{lipsum}
\usepackage{fancyhdr}       %
\usepackage{graphicx}       %
\graphicspath{{media/}}     %

\newcommand{\edit}[1]{\textcolor{black}{#1}}

\AtBeginDocument{%
  \providecommand\BibTeX{{%
    \normalfont B\kern-0.5em{\scshape i\kern-0.25em b}\kern-0.8em\TeX}}}

\title{T-RAG: Lessons from the LLM Trenches
}

\author{
  Masoomali Fatehkia, Ji Kim Lucas,   Sanjay Chawla\\
  Qatar Computing Research Institute \\
  Hamad Bin Khalifa University \\
  Doha\\
  \texttt{\{mfatehkia, jlucas, schawla\}@hbku.edu.qa} \\
}

\begin{document}
\maketitle

\begin{abstract}

Large Language Models (LLM) have shown remarkable language capabilities fueling attempts to integrate them into applications across a wide range of domains. An important application area is question answering over private enterprise documents where the main considerations are data security, which necessitates applications that can be deployed on-prem, limited computational resources and the need for a robust application that correctly responds to queries. Retrieval-Augmented Generation (RAG) has emerged as the most prominent framework for building LLM-based applications. While building a RAG is relatively straightforward, making it robust and a reliable application requires extensive customization and relatively deep knowledge of the application domain. We share our experiences building and deploying an LLM application for question answering over private organizational documents. Our application combines the use of RAG with a finetuned open-source LLM. 
Additionally, our system, which we call Tree-RAG (T-RAG), uses a tree structure to represent entity hierarchies within the organization. This is used to generate a textual description to augment the context when responding to user queries pertaining to entities within the organization's hierarchy. \edit{Our evaluations, including a Needle in a Haystack test,} show that this combination performs better than a simple RAG or finetuning implementation. 
Finally, we share some lessons learned based on our experiences building an LLM application for real-world use.

\end{abstract}

\maketitle

\section{Introduction}

Large Language Models (LLM) represent the most recent advances in Natural Language Processing (NLP) demonstrating a wide range of capabilities in language processing~\cite{zhao_survey_2023}. They came into prominence after ChatGPT, an application by OpenAI that opened for public testing, went viral\footnote{\url{https://www.nytimes.com/2022/12/05/technology/chatgpt-ai-twitter.html}}. This has fueled attempts to use LLMs for a variety of applications ranging from creative writing~\cite{gomez-rodriguez_confederacy_2023}, to programming~\cite{liventsev_fully_2023}, legal~\cite{louis_interpretable_2023} and medical~\cite{he_survey_2023} domains which require greater factual accuracy. %

A promising area of application for LLMs is question answering over proprietary organizational documents such as governance/policy manuals. Such documents are often a regular point of reference as they guide the day-to-day operations and decision making within an organization. This results in frequent references to such documents or to experts within the organization who respond to queries about such information. Hence there is potential for increased efficiency from having an application that can respond to a diverse range of user queries based on organizational documents. %

There are several considerations when deploying an LLM application in such settings. One major concern is the security risks given the confidential nature of such documents. As a result, it is not possible to use proprietary LLM models over an API due to data leakage risks\footnote{\url{https://mashable.com/article/ samsung-chatgpt-leak-details}}. This necessitates the use of open source models that can be deployed on-premise. %
A second concern is limited computational resources as well as relatively smaller training datasets that can be generated based on the available documents. Finally, any such application must be able to reliably and correctly respond to user queries. Therefore, deploying a robust application in such settings is not trivial, requiring many decisions and customization.%

\begin{oframed}
    \textbf{Use Case:}
    
    Our use case is question answering based on an organization's governance manual. The main features of such a document are (i) descriptions of the organization's governing principles, duties and responsibilities of various governing bodies and (ii) details about the full hierarchy of entities under the organization and their categorizations. An LLM application responding to questions based on the document should be able to answer a range of questions such as describing the various governing bodies, their responsibilities, as well as listing entities within the organization and the category they belong to. Below are some illustrative examples of the types of queries that a user a might ask based on a document about the UN organization: 

    \begin{itemize}
        \item How does Giga\footnotemark plan to involve UNICEF and ITU %
        counterparts in their strategy?
        \item Give examples of entities under HR Management?
        \item What are the three broad categories of audiences that Giga will target in 2023?
    \end{itemize}
\end{oframed}
\footnotetext{Giga is an initiative by UNICEF \& ITU that aims to connect every school to the internet. ITU is the International Telecommunication Union.}

In this work, we share our experiences building and deploying an LLM application for question answering over a private governance manual document for a large non-profit organization.  
We make the following contributions:

\begin{itemize}
    \item We present a real case study in creating an LLM powered application for question answering over the governance document for end users from an organization.
    \item We create an application that combines the use of Retrieval-Augmented Generation (RAG) with a finetuned open-source LLM for response generation, trained over an instruction dataset generated from the organization's document.
    \item We include a novel tree-based context as a component in our system which we call Tree-RAG (T-RAG). It uses a tree structure to represent hierarchical information i.e. entities in an organization. This is used to generate a textual description to augment the context when responding to user queries pertaining to entities within the organization's hierarchy. %
    \item We present a new evaluation metric (Correct-Verbose) for assessing the generated responses.  %
    This metric captures responses that are correct but which also provide additional correct information not relevant to the question. \edit{In addition, we perform a Needle in a Haystack test to evaluate the retrieval capability of T-RAG.}
\end{itemize}

The rest of the paper is organized as follows. Section~\ref{sec:rel_work} provides a review of the related literature. Section~\ref{sec:relevant_terms} defines some relevant terminology. Section~\ref{sec:rag} provides an overview of Retrieval Augmented Generation (RAG) for LLM applications and demonstrates T-RAG. Section~\ref{sec:methods} gives details about our system's implementation. Section~\ref{sec:results} presents evaluations of our system and section~\ref{sec:discuss} concludes the paper with a discussion and directions for future work. %

\section{Related Work}
\label{sec:rel_work}

\subsection{Large Language Models}
Large Language Models (LLM) have shown remarkable capabilities in Natural Language Processing~\cite{zhao_survey_2023}. Recent years have seen an explosion of different LLMs with examples including OpenAI's GPT series such as GPT-4~\cite{openai_gpt-4_2023} and open source models such as Meta's Llama-2~\cite{touvron_llama_2023}. LLMs are based on the transformer architecture~\cite{vaswani_attention_2017} with larger models having as many as hundreds of billions of parameters. They are trained on massive corpus of training data including books, crawled web pages and conversations from social media platforms~\cite{zhao_survey_2023}. 
Their language abilities make LLMs suitable for downstream applications such as question answering. However, LLMs face limitations in handling domain-specific or highly specialized queries that require information outside their training corpus~\cite{kandpal_large_2023}. LLMs can be pre-trained for specific domains such as finance~\cite{huang_finbert_2023} or geographic-language for mapping applications~\cite{huang_ernie-geol_2022}, but this requires a large training dataset and expensive computational resources. A variety of approaches have evolved for building domain-specific applications with LLMs, which we review here.

\subsection{Finetuning}

Finetuning is a method of incorporating domain knowledge into an LLM's parametric memory by updating the model's weights through training on a domain-specific labeled dataset such as a questions and answers dataset for Q\&A applications~\cite{min_question_2017}. It allows one to benefit from an LLM's language capabilities while incorporating knowledge of the new task and adapting the LLM's writing style and tone~\cite{gao_retrieval-augmented_2024}. Finetuning requires the creation of a high-quality training dataset but this is still much smaller than the scale of data needed for pre-training~\cite{gao_retrieval-augmented_2024}.
While full finetuning (updating all parameters of a model)~\cite{howard_universal_2018} is computationally prohibitive, comparable performance can be achieved by updating a significantly smaller subset of the model's parameters~\cite{pmlr-v97-houlsby19a}. The rise of Parameter-Efficient Finetuning (PEFT) methods in recent years has significantly reduced the memory footprint and computational resources required for finetuning~\cite{xu_parameter-efficient_2023,lialin_scaling_2023} making this a more accessible option for organizations with smaller resources.

\subsection{Retrieval-Augmented Generation (RAG)}

A popular apporach for building LLM applications that does not require training the LLM, is Retrieval Augmented Generation (RAG). 
When asked domain-specific questions outside their training data, LLMs can generate incorrect information or `hallucinations' \cite{zhang_sirens_2023}. RAG addresses this limitation by retrieving information from an external data source which is then passed as contextual information to the LLM model for response generation~\cite{lewis_retrieval-augmented_2020}. This results in improved factual accuracy and relevance of the generated responses by enabling the model to access external information sources~\cite{ram_-context_2023}. While RAG can be used during pre-training~\cite{guu_retrieval_2020,guu_realm_2020}, it is widely used during inference due to its practicality and relative easy of use~\cite{gao_retrieval-augmented_2024}. 
However, RAG is sensitive to the composition of the retrieved documents used to create the context~\cite{cuconasu_power_2024} and therefore requires extensive customization to build an effective retrieval pipeline. RAG can also be combined with other approaches such as finetuning~\cite{balaguer_rag_2024}.

\subsection{Knowledge Graphs}

While RAG applications typically rely on a retriever to fetch relevant documents based on a user query, there can be other approaches for retrieving relevant context. One such approach relies on the use of knowledge graphs to generate the context based on input queries~\cite{agrawal_can_2023}. Knowledge graphs represent symbolic knowledge of real world facts as triples representing pairs of entities (nodes in the graph) and their relationship (edges in the graph). Relevant information can be extracted from a knowledge graph based on entities mentioned in a user's query and provided as context either in a raw format as triples~\cite{baek_knowledge-augmented_2023} or rewritten into textual statements~\cite{wu_retrieve-rewrite-answer_2023}. Domain-specific knowledge graphs have been used for question answering applications in disciplines such as medicine~\cite{xia_medconqa_2022}, finance~\cite{baldazzi_fine-tuning_2023} and %
education~\cite{agrawal_aiseckg_2023}. \edit{Another approach, GraphRAG, uses an LLM-derived knowledge graph based on source documents for retrieval and response generation ~\cite{edge_local_2024}. While knowledge graphs capture diverse sets of relationships between pairs of entities, they do not capture hierarchical information such as organizational structures which would more naturally be captured in a tree structure.}

\subsection{Applications of LLMs}

The use of LLMs has been explored in various domains such as education for generating exam questions~\cite{drori_human_2023}, recruitment and job recommendation~\cite{fang_recruitpro_2023}, news recommendation~\cite{xiao_training_2022}, for a range of healthcare applications~\cite{he_survey_2023}, medical question answering~\cite{guo_medical_2022}, querying patient health records~\cite{hamidi_evaluation_2023}, for assistive mental health tools~\cite{lai_psy-llm_2023}, legal question answering~\cite{louis_interpretable_2023} and IT support systems~\cite{yang_empower_2023}.

\section{Relevant Terminology}
\label{sec:relevant_terms}

Below is a short glossary of some LLM related terminology:

\begin{itemize}
    \item \textbf{Prompt:} Any text that is provided as input to an LLM which conditions the model's behavior and the generated output. It can consist of multiple elements including instructions, context, questions and examples, depending on the task. %
    \item \textbf{System Prompt:} A text instruction placed at the beginning of a prompt and depending on the model demarcated by special tags (eg: <<SYS>>). It contains instructions and sets the setting for what the LLM is expected to do. An example system prompt that we used is provided in Appendix~\ref{appx:system_prompt}.%
    \item \textbf{Context:} Additional pieces of text added to a prompt that can help an LLM respond to a question, eg: for the question \emph{Where did the 2022 FIFA World Cup take place?} the context may be a relevant paragraph from a Wikipedia article. %
    \item \textbf{In-context Learning:} It is the ability of LLMs to perform a new task, without finetuning, by being given demonstrations in the context. Eg: to do sentiment analysis, we can give an LLM two sentences and their sentiment, then a third sentence, for which the model will output its prediction of the sentiment. This is also referred to as few-shot learning. %
    \item \textbf{Context Window/Length:} The maximum number of tokens that an LLM can take as input (4,096 tokens in Llama-2). A longer context window allows a model to process more information at once, useful for understanding longer texts.%
    \item \textbf{Hallucination:} When an LLM generates plausible but factually incorrect output that deviates from the context, user input or world knowledge~\cite{zhang_sirens_2023}. Eg: when asked to list entities under the \emph{UNHCR Innovation Service} (Figure~\ref{fig:tree_retrieval}), a model might plausibly but incorrectly mention \emph{Design Services} even though no such entity exists in the organization.
\end{itemize}

\section{Retrieval-Augmented Generation}%
\label{sec:rag}

Retrieval-Augmented Generation (RAG) enhances the performance of LLMs on domain specific tasks by providing the model with an external source of information. While there are many variations, we provide an overview of a typical RAG application in Algorithm 1. This generally consists of two processes, an Index process done once at the start of the application and the Query process which happens every time in response to incoming queries~\cite{barnett_seven_2024}. The index process occurs as follows. The input document \begin{math} D \end{math} is split into discrete chunks \begin{math} \{c_1, c_2, ..., c_n\}\end{math} (steps 2 \& 3). Using an encoder model, the split chunks \begin{math} c_i \end{math} are converted to embedding vectors \begin{math} \vec{d_i} = encoder(c_i) \end{math} (step 4) which are then stored in a vector database (step 5). This database is later used to retrieve relevant chunks for a given query.

The Query processing happens in response to incoming user queries. For a given query \begin{math} q \end{math}, the encoding model is used to create a vector embedding of the query \begin{math} \vec{v} = encoder(q) \end{math}. The database is then searched to find the top \begin{math} k \end{math} chunk embeddings \begin{math} \{\vec{d_1}, \vec{d_2}, ..., \vec{d_k}\}\end{math} that are similar to the query embedding \begin{math} \vec{v} \end{math}. There are various algorithms for determining similarity between the chunk embeddings \begin{math} \vec{d_i} \end{math}  and the query embedding \begin{math} \vec{v} \end{math} and how many and which chunks to fetch. The top \begin{math} k \end{math} chunks \begin{math} \{c_1, c_2, ..., c_k\}\end{math} retrieved from the database, along with the query, are then passed into the prompt template. The completed prompt is then input to an LLM model which generates an output based on the provided information. This response is then returned to the user.

\begin{table}[h]
  \caption{Algorithm for an LLM application. On the left is the algorithm for a typical RAG application and on the right is the algorithm for our system (T-RAG). The parts highlighted in blue are where our system differs from a typical RAG application. We do not show the Index Process for T-RAG as it is similar to RAG.}
  \label{tab:rag_algorithms}
  \begin{tabular}{cc}
    \toprule
    \begin{minipage}[t]{0.49\textwidth} 
    \textbf{Algorithm 1} High-level overview of a typical RAG system 
    \end{minipage} & \begin{minipage}[t]{0.49\textwidth} 
    \textbf{Algorithm 2} Our system (T-RAG) 
    \end{minipage} \\
    \midrule
    \begin{minipage}[t]{0.49\textwidth} 
        \textbf{Index Process}
        \begin{algorithmic}[1]
        \STATE $embeddings \leftarrow load("embedding\_model")$
        
        \STATE $doc \leftarrow load("file\_name")$
        \STATE $c \leftarrow chunker.chunk(doc)$
        \STATE $ce \leftarrow embeddings.embed(c)$
        \STATE $db \leftarrow index(ce)$
        
        \end{algorithmic}

        \textbf{Query Process}
        \begin{algorithmic}[1]
        \STATE INITIALIZE $sys\_prompt \leftarrow "You\ are\ an\ AI.... "$
        \STATE INITIALIZE $model \leftarrow load("llm\_model")$
        
        \WHILE{TRUE}
        \STATE $q \leftarrow get.user\_query()$
        \STATE $qe \leftarrow embeddings.embed(q)$
        \STATE $chunks \leftarrow db.search(qe)$
        \STATE $context \leftarrow merge(chunks)$
        \STATE $prompt \leftarrow create\_prompt(sys\_prompt, q, context)$
        \STATE $answer \leftarrow model.generate(prompt)$
        \ENDWHILE
        \end{algorithmic}
        
    \end{minipage} &
    \begin{minipage}[t]{0.49\textwidth}

    \textbf{Query Process}
    \begin{algorithmic}[1]
    \STATE INITIALIZE $sys\_prompt \leftarrow "You\ are\ an\ AI.... "$
    \STATE \textcolor{blue}{INITIALIZE $model \leftarrow load("finetund\_llm")$}
    
    \STATE \textcolor{blue}{INITIALIZE $tree \leftarrow build\_tree("entities\_file")$}
    
    \WHILE{TRUE}
    \STATE $q \leftarrow get.user\_query()$
    \STATE $qe \leftarrow embeddings.embed(q)$
    \STATE $chunks \leftarrow db.search(qe)$
    
    \STATE \textcolor{blue}{$entities \leftarrow parse\_entities(q)$} 
    \IF{\textcolor{blue}{entities $\neq$ null }}
        \STATE \textcolor{blue}{$entities\_info \leftarrow tree.search(entities)$}
        \STATE \textcolor{blue}{$context \leftarrow merge(entities\_info, chunks)$}
    \ELSE
        \STATE \textcolor{blue}{$context \leftarrow merge(chunks)$}
    \ENDIF

    \STATE $prompt \leftarrow create\_prompt(sys\_prompt, q, context)$
    
    \STATE $answer \leftarrow model.generate(prompt)$
    \ENDWHILE
    \end{algorithmic}

    \end{minipage} \\
    \bottomrule
  \end{tabular}
\end{table}

\subsection{T-RAG}

\begin{figure}[h]
  \centering
  \includegraphics[width=0.8\textwidth]{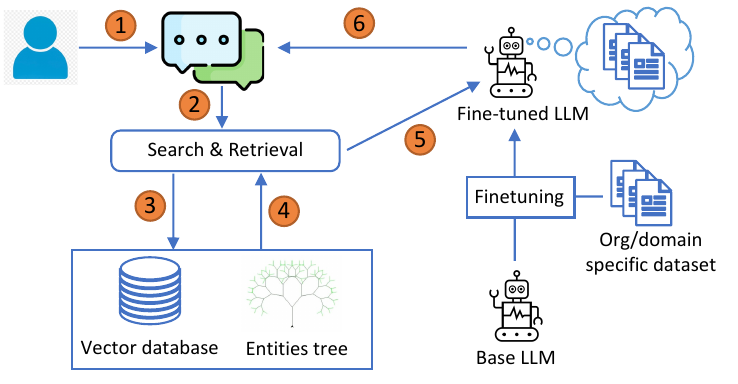}%
  \caption{Workflow of Tree-RAG (T-RAG). For a given user query, we search the vector database for relevant document chunks to be used as context. Additionally, if the query mentions entities from the organization, information about them is extracted from the entities tree and added to the context. We finetuned the Llama-2 7B model on an instruction dataset generated from the organization's document. We use the finetuned model for response generation.}
  \label{fig:system_chart}
\end{figure}

The overall workflow of our system, Tree-RAG (T-RAG), is shown in Figure~\ref{fig:system_chart} and outlined in Algorithm 2. %
Our system differs from the typical RAG application in the Query process. Instead of using an existing pre-trained LLM, we use a finetuned version of the LLM for answer generation; we finetuned the LLM model on an instruction dataset of questions and answers generated based on the organization's document as described in later sections. 

A feature of T-RAG is the inclusion of an entities tree in addition to the vector database for context retrieval. The entities tree holds information about entities in the organization and their location within the hierarchy. Each node in this tree represents an entity with the parent node indicating the group it belongs to. For example, in the UNHCR organizational structure shown in Figure~\ref{fig:tree_retrieval}, \emph{UNHCR Innovation Service} is an entity falling under the \emph{Deputy High Commissioner}. 

During retrieval, we use the entities tree to further augment the context retrieved by the vector database. %
The entity tree search and context generation occurs as follows. %
A parser module searches the user query for keywords matching the names of entities in the organization. If one or more matches are found, information about each matched entity is extracted from the tree and converted into a textual statement providing information about the entity and its location within the organization's hierarchy. This information is then combined with the document chunks retrieved from the vector database to form the context. This allows the model to access information about entities and their location within the organization's hierarchy when users ask questions about these entities.

\section{Methods}
\label{sec:methods}

In this section and subsequent ones, we will share details about our system and evaluations based on the organization's document. As we are not able to share specific details from this document, we will demonstrate our examples using publicly available UN organization document.

\subsection{Instruction Dataset Preparation}

Finetuning an LLM model requires a domain specific training dataset. Here, we describe the procedure we followed to generate an instruction dataset from the organization's document. 

The first step was to parse the original PDF document file into text format for further processing; this was done using the LangChain library\footnote{\url{https://python.langchain.com}}. 
In addition to text, the file also contained several tables and an image illustrating all entities in the organizational chart. The tables were manually converted to text by a human expert who wrote sentences describing the information in the tables. The organizational chart was converted to text in the same manner. 

The next step in the process was to divide the document into chunks. This was done based on the section headers in the document, splitting each section into a separate chunk. We then generated (question, answer) pairs for each chunk over several iterations, as follows. In the first iteration, for each chunk we prompted the Llama-2 model to generate questions and answers for the provided chunk. The model was prompted to produce a variety of question types such as True or False, Summary, Short Answer etc. The model responses providing the questions, answers and the relevant chunk were recorded. In the second iteration, for each chunk, the model was prompted with an example of a dialog and asked to produce a dialog between a user and AI assistant. In the third iteration, the model was asked to perform the same task (produce questions and answers for a given chunk); in this iteration the model was provided with examples of questions that were created by a human expert based on the document. The prompts are provided in appendix \ref{sec:instruction_dataset_prompts}. 

We aggregated the questions and answers produced by the various iterations to create our dataset. Quality checks were performed through manual inspection of the generated questions and answers. We also performed duplicate removal to remove any duplicated questions. Our dataset consists of 1,614 question and answer pairs that were randomly split into 90\% training and 10\% validation sets.

\subsection{LLM finetuning}

Full finetuning of LLMs is computationally expensive given their large number of parameters. Parameter-Efficient Fine-Tuning (PEFT) are a set of techniques for finetuning LLMs efficiently. One such technique is QLoRA \cite{dettmers_qlora_2023} which uses a combination of 4-bit Quantization of model weights and Low-Rank Adaptation (LoRA) \cite{hu_lora_2021} another technique for efficient finetuning of LLMs. 
The LoRA method is inspired by the hypothesis that the model weights in an LLM model have a low intrinsic rank making it possible to approximate them with low-rank matrices, significantly reducing the number of parameters that need to be updated during training. 

The LoRA method is represented in the equation below where \begin{math} h \end{math} is the model's output, \begin{math} x \end{math} is the input and \begin{math} W_0 \in \mathbb{R}^{d \times n} \end{math} are the pre-trained model weights. \begin{math} B \in \mathbb{R}^{d \times r}, A \in \mathbb{R}^{r \times n} \end{math} are a set of low-rank matrices (rank \begin{math} r \end{math} where \begin{math} r \ll \min(n, d) \end{math}) that are updated during finetuning while the pre-trained weights \begin{math} W_0 \end{math} are kept frozen. Hence, the number of parameters to be updated is \begin{math} d \times r + r \times n \end{math} which is much smaller than the total number of pre-traind parameters \begin{math} d \times n \end{math}. When finetuning Llama-2 7B, we used a rank \begin{math} r = 64 \end{math} resulting in about 33.5M trainable parameters which is a factor of about 200 times reduction compared to the full set of parameters in the model.
\begin{displaymath}
  h = W_0 x + BAx
\end{displaymath}

QLoRA achieves significant memory savings through weight quantization whereby the model weights \begin{math} W_0 \end{math} are reduced to a lower precision 4-bit representation.
We finetuned the base LLM model on our Q\&A instruction dataset with QLoRA, using the Hugging Face `peft'\footnote{\url{https://huggingface.co/docs/peft/index}} library. Finetuning was done on 4 Quadro RTX 6000 GPUs with a memory of 24GB.

\subsection{Tree Graph for Entities}

\begin{figure}[h]
  \centering
  \includegraphics[width=0.8\textwidth]{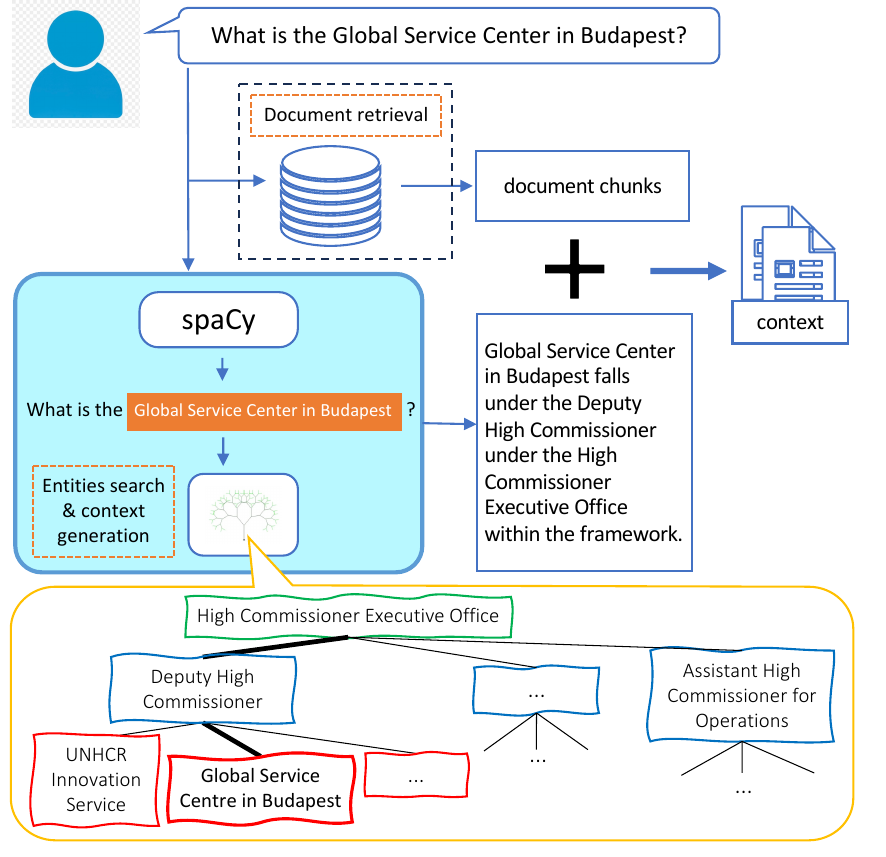}%
  \caption{Retrieval process for context generation. Here we use an illustrative example from a UNHCR organizational chart to show how the tree search and retrieval is done. In addition to retrieving contextual documents, we use the spaCy library with custom rules to detect named entities from the organization. If the query contains one or more such entities, then information about the location of that entity within the hierarchy is retrieved from the tree and formatted into textual statements; these are then added to the context in addition to the retrieved documents. If there are no mentioned entities in the user's query, then the tree search is skipped and only the context from the retrieved documents is used.}
  \label{fig:tree_retrieval}
\end{figure}

The document contained an organizational chart displaying the hierarchies and divisions within the organization. This included a listing of all entities under the organization and which particular categories and sub-categories they belonged to. In order to respond to queries regarding these entities and their place within the hierarchy, we need to represent this information in a format that can be accessed and used by the LLM model. This section explains our approach for representing this information and retrieving it when necessary, based upon user queries, to further enhance the context that is passed to the LLM model.

The organizational hierarchy and all entities within it are encoded in the form of a tree where each node represents some entity within the organization. The parent of each node represents the immediate category it belongs to. As an illustrative example, figure~\ref{fig:tree_retrieval} shows a snippet of the UNHCR\footnote{\url{https://reporting.unhcr.org/unhcr-headquarters-organizational-structure}} organizational chart. In the example shown, \emph{Global Service Center in Budapest} is an entity under the \emph{Deputy High commissioner} which is itself under the \emph{High Commissioner Executive Office}.
In this manner, the tree encodes the full hierarchy within the organization for each entity and can be used to trace the full path from the entity to the higher level categories it belongs to as well as to other entities that may fall under it. This information can be extracted during the retrieval step and converted into textual statements. These statements are then included in the context in addition to the document chunks retrieved from the vector database (see Figure~\ref{fig:tree_retrieval}). This enhances the context with relevant information about entities which can be used by the LLM model when generating a response.

In order to provide only the most relevant entity related information in the context, we need to detect whether and which of the organization's entities are mentioned in a user's query. If a user's query makes no reference to any entities from the organization, then the tree search is skipped and only the context from the retrieved documents is used. To enable such adaptive behavior, we need a way of detecting only named entities relevant to the organization. We used the spaCy\footnote{\url{https://spacy.io/}} library for detecting and extracting such named entities from the user's query. While the library has implemented a variety of algorithms for Named Entity Recognition (NER), using it out of the box will not work for custom use cases such as ours. For example, for the user's query in Figure~\ref{fig:tree_retrieval}, the spaCy library would detect \emph{Budapest} as a location entity mentioned in the text, missing the fact that \emph{Global Service Center in Budapest} is an entity within the UNHCR organization. Therefore, we customized the library for our use case by defining a new category with rules for detecting entities belonging to the organization using string matching. %

In summary, the overall workflow for context generation is shown in Figure~\ref{fig:tree_retrieval}. For the input user query, we first perform document retrieval from the vector database to identify relevant information chunks that should go into the context; this is standard procedure for any RAG application. However, we further augment the context by including additional information about entities within the organization based on the user's query. This is done by first parsing entity names from the user query. If there are no entities detected then the tree search is skipped and no further information is added to the context. However, if there are one or more entities mentioned by the user (the \emph{Global Service Center in Budapest} in our example), then the relevant information is extracted from the tree and converted into textual form. This information is then combined with the retrieved document chunks and used as the context.%

\subsection{Implementation configurations}

\subsubsection{LLM Model for Answer Generation}

A major requirement for our use case was deploying an application on-premise. As a result, we opted to use the Llama-2 model as it is an open-source model that achieves competitive performance~\cite{touvron_llama_2023}.
Llama-2 models are available in a range of parameter sizes from 7B to 70B. We used the smaller Llama-2 7B chat model\footnote{\url{https://huggingface.co/TheBloke/Llama-2-7B-Chat-GGML}} for our use case. Given the computational resources needed for finetuning and running large LLM models at runtime, smaller models are better suited for use by small and medium sized enterprises with limited computational resources or in geographic regions where access to GPUs are restricted~\cite{nellis_us_2023}.

\subsubsection{System Implementations}

We tried several setups for our system. These are summarized in Table~\ref{tab:model_version} and described in more detail here. %
The first implementation was RAG which used the base Llama-2 model for generating answers using chunks from the original document as context; we used a context size of 10 document chunks. The next implementation was to use a finetuned model with no contextual information. The Llama-2 model was finetuned on the instruction dataset that was generated from the organization's document. %

Finally, we implemented T-RAG, which combined the use of RAG to retrieve relevant context with the use of the finetuned model for response generation. T-RAG uses contextual information from two sources: (i) the pairs of questions and answers from the instruction dataset that were generated from the original document (we only kept the direct question and answers and filtered out other question types such as dialog, fill in the blank etc. we used a context size of 3 Q\&A pairs), and (ii) tree context which provides information, extracted from the entities tree as explained previously, for the entities mentioned in the user's query. %

\begin{table}[h]
  \caption{An overview of various system implementations}
  \centering
  \label{tab:model_version}
  \begin{tabular}{lcc}
    \toprule
    Name & Model & Context\\
    \midrule
    RAG & base Llama-2 & document chunks \\
    Finetuned & Finetuned Llama-2 & - \\
    T-RAG & Finetuned Llama-2 & Q\&A chunks + entities\\
  \bottomrule
\end{tabular}
\end{table}

We used the Chroma DB\footnote{\url{https://www.trychroma.com/}} vector database for storing the documents chunks for context retrieval. We used Maximum Marginal Relevance (MMR) for document selection during retrieval; this algorithm selects documents based on a combination of similarity to the input query while also optimizing for diversity of retrieved documents. For the embedding model we used `Instructor', a text embedding model that can produce embeddings for a variety of domains~\cite{su-etal-2023-one}. During inference, we use greedy decoding (temperature is 0) with a repetition penalty of 1.1 to generate responses. %

\section{Results}
\label{sec:results}

\subsection{Performance Evaluation}

Evaluating the outputs of LLMs is commonly done using either automated or human evaluation~\cite{chang_survey_2023}. Automated evaluations rely on the use of another larger, more powerful LLM such as GPT-4~\cite{liu_g-eval_2023} or a judge LLM tuned for the evaluation task~\cite{wang_pandalm_2023}. %
Studies have found LLM evaluations to be consistent with human evaluations for various tasks~\cite{chiang_can_2023}. While automated evaluations scale more easily and may be cheaper to produce, feedback from real users is critical for systems that will ultimately be used by humans. %

We relied on human evaluations to check the performance of our system. The system underwent three rounds of testing with end users from the organization. This resulted in three sets of questions that we used in our evaluations. The first set was curated by a human expert from the organization familiar with the document and was used for the initial testing of the system. The second and third sets of questions were generated during additional rounds of testing by other end users from the organization. 

The generated responses were evaluated by a human expert who marked the response as being either (i) Correct (C) if the response answered the question and was factually correct, or (ii) Correct-Verbose (CV) if the response answered the question but also provided other factually correct information not relevant to the question. %
\edit{An important goal for an LLM application is providing correct, concise answers. When generating a response, the LLM might pick up additional irrelevant information from the provided context. The motivation behind the CV metric is to quantify this by providing a measure of how often the system provides verbose answers by including additional factually correct but irrelevant information.} 
The results are reported in Table~\ref{tab:eval_results}. The aggregated results across all three sets is shown in bold under the label `All' in the `Questions Set' column. Column (T) in the table is the total number of responses that were either Correct or Correct-Verbose.

Across all 37 questions, RAG and Finetuned achieved similar performance, correctly answering 21 and 20 questions respectively. %
T-RAG achieved better performance overall, answering 27 questions correctly out of 37 total questions. However, T-RAG was also more prone to providing verbose answers; 6 of the answers provided by T-RAG were `Correct-Verbose' compared to only one question for the other implementations. 

\begin{table}[h]
  \caption{Evaluation Results: Each system was tested on three sets of questions generated from several rounds of user testing. The aggregated results from all three sets is shown under `All'. N is the number of questions in each set. The answers were scored manually as being Correct (C) or Correct-Verbose (CV) if in addition to being correct the answer provided additional correct information not relevant to the question. `T' is the total number of correct responses (T = C + CV) with the percentage shown in the last column. RAG and Finetuned performed similarly. T-RAG achieved better performance overall, but it was also more prone to verbose answers. }
  \centering
  \label{tab:eval_results}
    \begin{tabular}{lcccccc}
    \toprule 
    Name & Question Set & N & C & CV & T & Perc. \\
    \midrule

    RAG & set 1 & 17 & 9 & 0 & 9 & 52.9\% \\
    RAG & set 2 & 11 & 7 & 0 & 7 & 63.6\% \\
    RAG & set 3 & 9 & 4 & 1 & 5 & 55.6\% \\
    RAG & \textbf{All} & \textbf{37} & \textbf{20} & \textbf{1} & \textbf{21} & \textbf{56.8\%} \\
    \hline
    Finetuned & set 1 & 17 & 11 & 1 & 12 & 70.6\% \\
    Finetuned & set 2 & 11 & 3 & 0 & 3 & 27.3\% \\
    Finetuned & set 3 & 9 & 5 & 0 & 5 & 55.6\% \\
    Finetuned & \textbf{All} & \textbf{37} & \textbf{19} & \textbf{1} & \textbf{20} & \textbf{54.1\%} \\
    \hline
    T-RAG & set 1 & 17 & 9 & 4 & 13 & 76.5\% \\
    T-RAG & set 2 & 11 & 6 & 2 & 8 & 72.7\% \\
    T-RAG & set 3 & 9 & 6 & 0 & 6 & 66.7\% \\
    T-RAG & \textbf{All} & \textbf{37} & \textbf{21} & \textbf{6} & \textbf{27} & \textbf{73.0\%} \\

    \bottomrule
\end{tabular}
\end{table}

\subsection{Evaluating The Entity Tree Search Module}

In order to evaluate the performance benefits of the context generated by the tree component in T-RAG, we created two sets of entity-related questions. We name these test sets (i) simple and (ii) complex. The simple set consists of direct questions asking about an entity in the organization. %
The complex set, on the other hand, contains questions asking for a list of some or all entities falling under a category, or compound questions asking about entities falling under two different categories. As an illustrative example, based on the UNHCR organizational chart, a simple question would be \emph{Where is the Deputy High Commissioner in the framework?}, while a complex question would be \emph{Names some entities under Deputy high Commissioner and some under External Relations?}.

We tested different implementations with and without the tree context. As can be seen in Table~\ref{tab:tree_eval_res}, including the tree context improved the accuracy of the answers for both simple and complex questions. Including the tree context roughly doubled the number of correct responses generated by the Finetuned model on both sets of questions. This improvement can be attributed to several reasons. One observation is that including the tree context reduces model hallucinations whereby the model makes up non-existent entities or  categories. %
Another explanation is that the context augments the finetuned model's memory so it can provide more accurate responses (such as listing entities falling under a category), rather than relying purely on its parametric memory, as LLMs may struggle with remembering long-tail knowledge about less popular entities \cite{kandpal_large_2023}.

We see a similar, though slightly smaller, effect with T-RAG when the tree context is excluded. The improvement was modest for simple questions as T-RAG without tree got 16 out of 17 questions correctly. T-RAG without tree performed well on the direct questions as it had access to the contextual information from the Q\&A instruction dataset. The benefit of the tree context can be seen for the complex questions where there was a much larger improvement: out of 22 complex questions, T-RAG without tree answered 10 correctly versus 15 when the tree context was included. %

\begin{table}
  \caption{Evaluation results for the tree context for entity related questions. We compare several implementations with and without the tree context, in order to assess the performance improvements resulting from it. The columns show the number of correct answers with the percentage indicated in brackets. The tree context significantly improved performance over RAG and Finetuned.}
  \centering
  \label{tab:tree_eval_res}
  \begin{tabular}{lcc}
    \toprule

    Name & \begin{minipage}[c]{0.15\columnwidth} Simple \\ (17 questions)\end{minipage} & \begin{minipage}[c]{0.15\columnwidth} Complex \\ (22  questions)\end{minipage} \\
    \midrule
    RAG & 10 (58.5\%) & 11 (50\%)\\ 
    Finetuned & 8 (47.1\%) &  8 (36.4\%) \\
    Finetuned + tree & 17 (100\%) & 17 (77.3\%)  \\
    T-RAG without tree & 16 (94.1\%) & 10 (45.5\%) \\
    T-RAG & 17 (100\%) & 15 (68.2\%)\\
    
    \bottomrule
  \end{tabular}
\end{table}

\subsubsection{Evaluation on Entities from UNHCR Organizational Chart}

We also tested the effect of the tree generated context using UNHCR's organizational structure. We created a set of 13 simple and 13 complex questions asking about various entities within UNHCR's organizational hierarchy. We compared the performance of a RAG system using as context (i) document chunks from a text explaining UNHCR's organizational hierarchy vs. (ii) context generated from the organization's entity tree. 
As before, we observed improved accuracy when using the tree context. Of the 13 simple questions RAG with document chunks answered 6 (46.2\%) correctly while RAG with tree answered 12 (92.3\%). For the 13 complex questions RAG with document chunks answered 6 (46.2\%) correctly while RAG with tree answered 8 (61.5\%) correctly.

\subsubsection{Qualitative Evaluations}

Looking at the answers generated by the models we notice the following types of errors that are reduced from having the tree context. One source of error mentioned earlier that was mostly observed with the finetuned model (without any context) were hallucinations. %
Another source of error was misaligned answers whereby the response mentioned entities from the organization but listed them under a wrong category. Table~\ref{tab:tree_qual_eval_unhcr} shows an example response based on the UNHCR organizational chart illustrating this type of error. The question asks about entities under the \emph{Deputy High Commissioner}, however, the response incorrectly lists entities from the \emph{High Commissioner Executive Office}.

\begin{table}[h]
  \caption{Misalignment error: An example question and response generated by RAG with document chunks (no tree context) based on UNHCR entities. The entities in bold-red are actually under the \emph{High Commissioner Executive Office} but incorrectly listed under the \emph{Deputy High Commissioner}.}
  \centering
  \label{tab:tree_qual_eval_unhcr}
  \begin{tabular}{ll}
    \toprule
    Question & \begin{minipage}[t]{0.5\columnwidth} Name some entities under Deputy high Commissioner and some under External Relations?\end{minipage} \\[0.4cm]
    \midrule

     Response & \begin{minipage}[t]{0.5\columnwidth} Based on the provided context, the entities under the Deputy High Commissioner include:

\textcolor{red}{\textbf{1. Deputy High Commissioner}}

\textcolor{red}{\textbf{2. Inspector General Office}}

\textcolor{red}{\textbf{3. Policy Development \& Evaluation Office}}

And the entities under External Relations include:

1. Governance \& Partnership

2. Inter-agency Coordination

3. Strategic Communications

4. Private Sector Fund Raising Service

5. Communications \& Public Information

\end{minipage}\\
    
    \bottomrule
  \end{tabular}
\end{table}

\subsubsection{\edit{Needle in a Haystack test}}

\edit{We performed a Needle in a Haystack test~\cite{LLMTest_NeedleInAHaystack} to further evaluate how the performance of the model is affected based on the position of the context. This is done by placing the relevant context (context retrieved from tree for T-RAG or the document chunks containing the relevant context for RAG) at different positions in the context surrounded by additional $k$ unrelated context chunks. The performance was then evaluated on all 39 entity related questions.   }

\edit{The results are presented in Table~\ref{tab:needle_in_haystack}. Generally, the model performs better when the relevant context is placed at the ends rather than at the middle. Notably, the performance of RAG decreased significantly (33.3\%) when the relevant context chunks were placed in the middle surrounded by other unrelated context ($k=10$). In comparison, T-RAG performed much better (64.1\%) even when the tree context was placed in the middle amid 10 other unrelated chunks. As these results suggest, the model generally performs better at retrieving the correct answer from the tree context even when it is placed in different positions with different context sizes. }

\begin{table}[h]
  \caption{\edit{Needle in a Haystack test. The tree context (or document chunks containing relevant context in the case of RAG) are placed at different positions in the context surrounded by $k$ other unrelated documents. The performance was evaluated on the 39 entity related questions. The models generally perform better when the relevant information is at the ends of the context than at the middle. RAG performance decreased significantly when the relevant document chunks were in the middle among many other unrelated documents ($k=10$) while T-RAG still performed better in this case. }}
  \centering
  \label{tab:needle_in_haystack}
    \begin{tabular}{lcccc}
    \toprule 
     & \multicolumn{2}{c}{T-RAG} & \multicolumn{2}{c}{RAG} \\ 
    Position & $k=2$ & $k=10$ & $k=2$ & $k=10$ \\
    
    \midrule

    Top & 29 (74.4\%) & 27 (69.2\%) & 30 (76.9\%) & 22 (56.4\%) \\

    Mid & 27 (69.2\%) & 25 (64.1\%) & 26 (66.7\%) & 13 (33.3\%) \\

    End & 32 (82.1\%) & 28 (71.8\%) & 25 (64.1\%) & 18 (46.2\%) \\
    
    \bottomrule
\end{tabular}
\end{table}

\subsection{Overfitting Test of Finetuned Model}

\begin{figure}[h]
  \centering
  \includegraphics[width=0.7\textwidth]{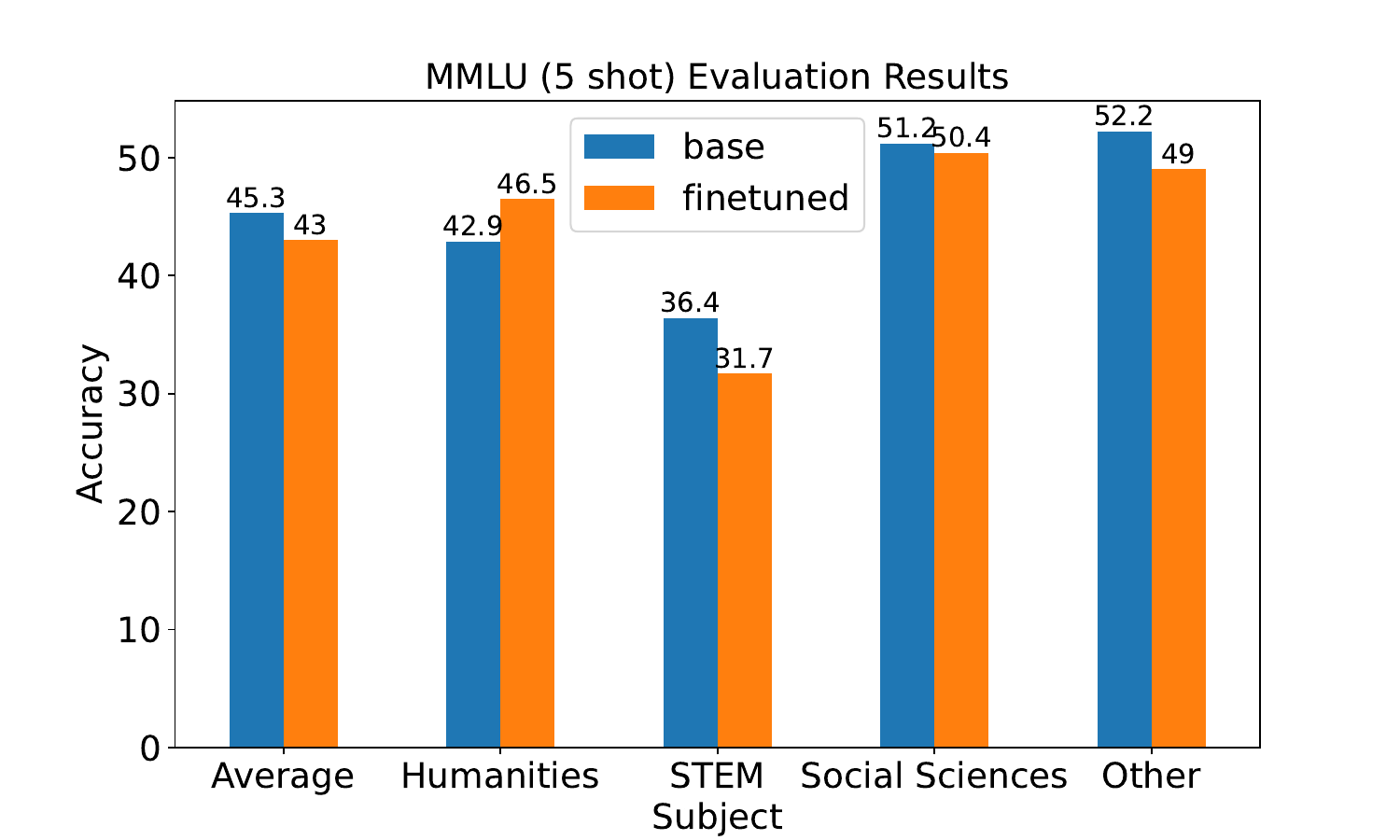}%
  \caption{Performance of the finetuned and base Llama-2 models on MMLU. %
  Both achieve similar overall accuracy, however, the finetuned model scores somewhat higher on the Humanities and lower on STEM. Finetuning should be done with caution as it can impact a model's general capabilities. %
  }
  \label{fig:mmlu_eval}
\end{figure}

Finetuning enables an LLM to learn new tasks by updating its weights. However, this poses a risk of the model overfitting the dataset, forgetting what it had learnt during pre-training. To test for this we compared the performance of our finetuned model to the base model on the Massive Multitask Language Understanding (MMLU) benchmark~\cite{hendrycks_measuring_2021}. MMLU is used to evaluate LLMs for language understanding and knowledge; it consisting of multiple-choice questions spanning a range of subjects such as STEM, the humanities, the social sciences and more. %
Figure~\ref{fig:mmlu_eval} shows the overall and subject specific accuracy of the base and finetuned models. The finetuned model achieves an overall accuracy of 43\% compared to 45.3\% for the base model. While finetuning does not appear to have led to overfitting here, it is important to be careful with finetuning as it can impact the general language abilities of an LLM.%

\section{Lessons Learned}
\label{sec:discuss}

There are significant considerations and customization that go into building a robust LLM system for practical applications. %
Here are some lessons we can share based on our experiences:
\begin{itemize}
    \item While building an initial RAG application is easy, making it robust is non-trivial, requiring domain knowledge expertise (we used the help of a domain expert to curate example questions that were used in our prompts to generate the instruction dataset for finetuning) and many design choices to optimize different components of the system.%
    \item Finetuned models can be sensitive to the phrasing of questions. %
    For example, when the finetuned model was asked to provide "a \emph{comprehensive} list of \emph{all} the..." vs. "a list of \emph{all} the...", the model's response to the former included hallucinations with made up names while the latter question was answered correctly. We observed this with other variations in question phrasing and hypothesize that differences in phrasing from the training dataset may be one explanation. %
    \item Finetuned models can save space on the limited context windows of LLMs by incorporating information into the model's parameters, thus reducing the required amount of context. %
    This can leave more room for other information such as conversation history for chat applications. We believe this can be a simpler alternative to ongoing efforts aimed at increasing the context window of LLMs~\cite{zhang_extending_2024} and adapting RAG to these larger context windows~\cite{xu_retrieval_2024}. %
    \item %
    Involving end users for testing at various phases of system development can generate feedback that can help steer some of the decision making during development. %
    \item Trees provide an appropriate structure for representing hierarchical information such as entities in an organization that can be used to enhance the context. Our evaluations suggest that this helps make the system fairly robust in responding to questions about entities.%
\end{itemize}

\subsection{RAG vs. Finetuning}

We used a combination of RAG with finetuning in our application. Both approaches have their strengths and weaknesses. Finetuning requires more computational resources initially to train a model. For smaller applications, the computational requirements of RAG are likely to be minimal beyond the retrieval technologies used. %
Compared to RAG, Finetuning can allow for adapting the model's writing style and tone to match the organization's document. However, finetuning must be done carefully as updating a model's parameters can degrade its overall language abilities. Testing finetuned models for overfitting, as we have done here, can be a useful way to check for this. %
While finetuning requires efforts to curate a high quality training dataset and tune hyper-parameters, RAG too requires many optimizations. Looking at the RAG algorithm presented in Table~\ref{tab:rag_algorithms}, there are many settings that can be explored at each step, including how to chunk the source document, choice of embedding model and retrieval algorithm among others. %

Maintaining LLM applications requires updating their knowledge bases as the underlying documents change over time. For RAG, this can be done readily by updating the retrieval database, making it more suitable for dynamic, frequent updates. Updating a finetuned model, however, requires preparing a training dataset and retraining the model, making it more suited to applications such as ours where the underlying document changes less frequently. %

Deployed applications receive a diverse variety of user queries. As we observed earlier, finetuned models are still prone to hallucinations especially when faced with unfamiliar input. Using RAG can reduce hallucinations considerably by grounding the model's responses in the given context~\cite{zhang_sirens_2023}. However, RAG also suffers from many limitations across the pipeline and is sensitive to noisy or incomplete contexts resulting in hallucinations and incomplete answers~\cite{barnett_seven_2024}. Reducing LLM hallucinations is still an open research question and future advances can help make systems more robust as frameworks mature over time~\cite{ye_cognitive_2023}. %
From our experience, hybrid approaches combining RAG and finetuning are likely to be promising for real-world applications and should be explored further.

\subsection{Future Work}

We received positive feedback on our system from the organization and there is interest to expand the current system to a wider corpus of documents. Another area for future work is to expand the system into a chat-based application. This requires further considerations beyond what is needed for a Q\&A application such as effectively handling chat history.

\bibliographystyle{ACM-Reference-Format}
\bibliography{sources}


\begin{thebibliography}{52}


\ifx \showCODEN    \undefined \def \showCODEN     #1{\unskip}     \fi
\ifx \showDOI      \undefined \def \showDOI       #1{#1}\fi
\ifx \showISBNx    \undefined \def \showISBNx     #1{\unskip}     \fi
\ifx \showISBNxiii \undefined \def \showISBNxiii  #1{\unskip}     \fi
\ifx \showISSN     \undefined \def \showISSN      #1{\unskip}     \fi
\ifx \showLCCN     \undefined \def \showLCCN      #1{\unskip}     \fi
\ifx \shownote     \undefined \def \shownote      #1{#1}          \fi
\ifx \showarticletitle \undefined \def \showarticletitle #1{#1}   \fi
\ifx \showURL      \undefined \def \showURL       {\relax}        \fi
\providecommand\bibfield[2]{#2}
\providecommand\bibinfo[2]{#2}
\providecommand\natexlab[1]{#1}
\providecommand\showeprint[2][]{arXiv:#2}

\bibitem[Agrawal et~al\mbox{.}(2023a)]%
        {agrawal_can_2023}
\bibfield{author}{\bibinfo{person}{Garima Agrawal}, \bibinfo{person}{Tharindu Kumarage}, \bibinfo{person}{Zeyad Alghami}, {and} \bibinfo{person}{Huan Liu}.} \bibinfo{year}{2023}\natexlab{a}.
\newblock \bibinfo{title}{Can {Knowledge} {Graphs} {Reduce} {Hallucinations} in {LLMs}? : {A} {Survey}}.
\newblock
\newblock
\urldef\tempurl%
\url{https://doi.org/10.48550/arXiv.2311.07914}
\showDOI{\tempurl}
\newblock
\shownote{arXiv:2311.07914 [cs]}.


\bibitem[Agrawal et~al\mbox{.}(2023b)]%
        {agrawal_aiseckg_2023}
\bibfield{author}{\bibinfo{person}{Garima Agrawal}, \bibinfo{person}{Kuntal Pal}, \bibinfo{person}{Yuli Deng}, \bibinfo{person}{Huan Liu}, {and} \bibinfo{person}{Chitta Baral}.} \bibinfo{year}{2023}\natexlab{b}.
\newblock \showarticletitle{{AISecKG}: {Knowledge} {Graph} {Dataset} for {Cybersecurity} {Education}}. In \bibinfo{booktitle}{\emph{Proceedings of the {AAAI} 2023 {Spring} {Symposium} on {Challenges} {Requiring} the {Combination} of {Machine} {Learning} and {Knowledge} {Engineering} ({AAAI}-{MAKE} 2023)}} \emph{(\bibinfo{series}{{CEUR} {Workshop} {Proceedings}}, Vol.~\bibinfo{volume}{3433})}, \bibfield{editor}{\bibinfo{person}{Andreas Martin}, \bibinfo{person}{Hans-Georg Fill}, \bibinfo{person}{Aurona Gerber}, \bibinfo{person}{Knut Hinkelmann}, \bibinfo{person}{Doug Lenat}, \bibinfo{person}{Reinhard Stolle}, {and} \bibinfo{person}{Frank~van Harmelen}} (Eds.). \bibinfo{publisher}{CEUR}, \bibinfo{address}{Hyatt Regency, San Francisco Airport}.
\newblock
\urldef\tempurl%
\url{https://ceur-ws.org/Vol-3433/#paper6}
\showURL{%
\tempurl}
\newblock
\shownote{ISSN: 1613-0073}.


\bibitem[Baek et~al\mbox{.}(2023)]%
        {baek_knowledge-augmented_2023}
\bibfield{author}{\bibinfo{person}{Jinheon Baek}, \bibinfo{person}{Alham~Fikri Aji}, {and} \bibinfo{person}{Amir Saffari}.} \bibinfo{year}{2023}\natexlab{}.
\newblock \showarticletitle{Knowledge-{Augmented} {Language} {Model} {Prompting} for {Zero}-{Shot} {Knowledge} {Graph} {Question} {Answering}}. In \bibinfo{booktitle}{\emph{Proceedings of the 1st {Workshop} on {Natural} {Language} {Reasoning} and {Structured} {Explanations} ({NLRSE})}}, \bibfield{editor}{\bibinfo{person}{Bhavana Dalvi~Mishra}, \bibinfo{person}{Greg Durrett}, \bibinfo{person}{Peter Jansen}, \bibinfo{person}{Danilo Neves~Ribeiro}, {and} \bibinfo{person}{Jason Wei}} (Eds.). \bibinfo{publisher}{Association for Computational Linguistics}, \bibinfo{address}{Toronto, Canada}, \bibinfo{pages}{78--106}.
\newblock
\urldef\tempurl%
\url{https://doi.org/10.18653/v1/2023.nlrse-1.7}
\showDOI{\tempurl}


\bibitem[Balaguer et~al\mbox{.}(2024)]%
        {balaguer_rag_2024}
\bibfield{author}{\bibinfo{person}{Angels Balaguer}, \bibinfo{person}{Vinamra Benara}, \bibinfo{person}{Renato Luiz de~Freitas Cunha}, \bibinfo{person}{Roberto de M.~Estevão Filho}, \bibinfo{person}{Todd Hendry}, \bibinfo{person}{Daniel Holstein}, \bibinfo{person}{Jennifer Marsman}, \bibinfo{person}{Nick Mecklenburg}, \bibinfo{person}{Sara Malvar}, \bibinfo{person}{Leonardo~O. Nunes}, \bibinfo{person}{Rafael Padilha}, \bibinfo{person}{Morris Sharp}, \bibinfo{person}{Bruno Silva}, \bibinfo{person}{Swati Sharma}, \bibinfo{person}{Vijay Aski}, {and} \bibinfo{person}{Ranveer Chandra}.} \bibinfo{year}{2024}\natexlab{}.
\newblock \bibinfo{title}{{RAG} vs {Fine}-tuning: {Pipelines}, {Tradeoffs}, and a {Case} {Study} on {Agriculture}}.
\newblock
\newblock
\urldef\tempurl%
\url{https://doi.org/10.48550/arXiv.2401.08406}
\showDOI{\tempurl}
\newblock
\shownote{arXiv:2401.08406 [cs]}.


\bibitem[Baldazzi et~al\mbox{.}(2023)]%
        {baldazzi_fine-tuning_2023}
\bibfield{author}{\bibinfo{person}{Teodoro Baldazzi}, \bibinfo{person}{Luigi Bellomarini}, \bibinfo{person}{Stefano Ceri}, \bibinfo{person}{Andrea Colombo}, \bibinfo{person}{Andrea Gentili}, {and} \bibinfo{person}{Emanuel Sallinger}.} \bibinfo{year}{2023}\natexlab{}.
\newblock \showarticletitle{Fine-{Tuning} {Large} {Enterprise} {Language} {Models} via {Ontological} {Reasoning}}. In \bibinfo{booktitle}{\emph{Rules and {Reasoning}: 7th {International} {Joint} {Conference}, {RuleML}+{RR} 2023, {Oslo}, {Norway}, {September} 18–20, 2023, {Proceedings}}}. \bibinfo{publisher}{Springer-Verlag}, \bibinfo{address}{Berlin, Heidelberg}, \bibinfo{pages}{86--94}.
\newblock
\showISBNx{978-3-031-45071-6}
\urldef\tempurl%
\url{https://doi.org/10.1007/978-3-031-45072-3_6}
\showDOI{\tempurl}


\bibitem[Barnett et~al\mbox{.}(2024)]%
        {barnett_seven_2024}
\bibfield{author}{\bibinfo{person}{Scott Barnett}, \bibinfo{person}{Stefanus Kurniawan}, \bibinfo{person}{Srikanth Thudumu}, \bibinfo{person}{Zach Brannelly}, {and} \bibinfo{person}{Mohamed Abdelrazek}.} \bibinfo{year}{2024}\natexlab{}.
\newblock \bibinfo{title}{Seven {Failure} {Points} {When} {Engineering} a {Retrieval} {Augmented} {Generation} {System}}.
\newblock
\newblock
\urldef\tempurl%
\url{http://arxiv.org/abs/2401.05856}
\showURL{%
\tempurl}
\newblock
\shownote{arXiv:2401.05856 [cs]}.


\bibitem[Chang et~al\mbox{.}(2023)]%
        {chang_survey_2023}
\bibfield{author}{\bibinfo{person}{Yupeng Chang}, \bibinfo{person}{Xu Wang}, \bibinfo{person}{Jindong Wang}, \bibinfo{person}{Yuan Wu}, \bibinfo{person}{Linyi Yang}, \bibinfo{person}{Kaijie Zhu}, \bibinfo{person}{Hao Chen}, \bibinfo{person}{Xiaoyuan Yi}, \bibinfo{person}{Cunxiang Wang}, \bibinfo{person}{Yidong Wang}, \bibinfo{person}{Wei Ye}, \bibinfo{person}{Yue Zhang}, \bibinfo{person}{Yi Chang}, \bibinfo{person}{Philip~S. Yu}, \bibinfo{person}{Qiang Yang}, {and} \bibinfo{person}{Xing Xie}.} \bibinfo{year}{2023}\natexlab{}.
\newblock \bibinfo{title}{A {Survey} on {Evaluation} of {Large} {Language} {Models}}.
\newblock
\newblock
\urldef\tempurl%
\url{https://doi.org/10.48550/arXiv.2307.03109}
\showDOI{\tempurl}
\newblock
\shownote{arXiv:2307.03109 [cs]}.


\bibitem[Chiang and Lee(2023)]%
        {chiang_can_2023}
\bibfield{author}{\bibinfo{person}{Cheng-Han Chiang} {and} \bibinfo{person}{Hung-yi Lee}.} \bibinfo{year}{2023}\natexlab{}.
\newblock \showarticletitle{Can {Large} {Language} {Models} {Be} an {Alternative} to {Human} {Evaluations}?}. In \bibinfo{booktitle}{\emph{Proceedings of the 61st {Annual} {Meeting} of the {Association} for {Computational} {Linguistics} ({Volume} 1: {Long} {Papers})}}, \bibfield{editor}{\bibinfo{person}{Anna Rogers}, \bibinfo{person}{Jordan Boyd-Graber}, {and} \bibinfo{person}{Naoaki Okazaki}} (Eds.). \bibinfo{publisher}{Association for Computational Linguistics}, \bibinfo{address}{Toronto, Canada}, \bibinfo{pages}{15607--15631}.
\newblock
\urldef\tempurl%
\url{https://doi.org/10.18653/v1/2023.acl-long.870}
\showDOI{\tempurl}


\bibitem[Cuconasu et~al\mbox{.}(2024)]%
        {cuconasu_power_2024}
\bibfield{author}{\bibinfo{person}{Florin Cuconasu}, \bibinfo{person}{Giovanni Trappolini}, \bibinfo{person}{Federico Siciliano}, \bibinfo{person}{Simone Filice}, \bibinfo{person}{Cesare Campagnano}, \bibinfo{person}{Yoelle Maarek}, \bibinfo{person}{Nicola Tonellotto}, {and} \bibinfo{person}{Fabrizio Silvestri}.} \bibinfo{year}{2024}\natexlab{}.
\newblock \bibinfo{title}{The {Power} of {Noise}: {Redefining} {Retrieval} for {RAG} {Systems}}.
\newblock
\newblock
\urldef\tempurl%
\url{https://doi.org/10.48550/arXiv.2401.14887}
\showDOI{\tempurl}
\newblock
\shownote{arXiv:2401.14887 [cs]}.


\bibitem[Dettmers et~al\mbox{.}(2023)]%
        {dettmers_qlora_2023}
\bibfield{author}{\bibinfo{person}{Tim Dettmers}, \bibinfo{person}{Artidoro Pagnoni}, \bibinfo{person}{Ari Holtzman}, {and} \bibinfo{person}{Luke Zettlemoyer}.} \bibinfo{year}{2023}\natexlab{}.
\newblock \bibinfo{title}{{QLoRA}: {Efficient} {Finetuning} of {Quantized} {LLMs}}.
\newblock
\newblock
\urldef\tempurl%
\url{https://doi.org/10.48550/arXiv.2305.14314}
\showDOI{\tempurl}
\newblock
\shownote{arXiv:2305.14314 [cs]}.


\bibitem[Drori et~al\mbox{.}(2023)]%
        {drori_human_2023}
\bibfield{author}{\bibinfo{person}{Iddo Drori}, \bibinfo{person}{Sarah~J. Zhang}, \bibinfo{person}{Reece Shuttleworth}, \bibinfo{person}{Sarah Zhang}, \bibinfo{person}{Keith Tyser}, \bibinfo{person}{Zad Chin}, \bibinfo{person}{Pedro Lantigua}, \bibinfo{person}{Saisamrit Surbehera}, \bibinfo{person}{Gregory Hunter}, \bibinfo{person}{Derek Austin}, \bibinfo{person}{Leonard Tang}, \bibinfo{person}{Yann Hicke}, \bibinfo{person}{Sage Simhon}, \bibinfo{person}{Sathwik Karnik}, \bibinfo{person}{Darnell Granberry}, {and} \bibinfo{person}{Madeleine Udell}.} \bibinfo{year}{2023}\natexlab{}.
\newblock \showarticletitle{From {Human} {Days} to {Machine} {Seconds}: {Automatically} {Answering} and {Generating} {Machine} {Learning} {Final} {Exams}}. In \bibinfo{booktitle}{\emph{Proceedings of the 29th {ACM} {SIGKDD} {Conference} on {Knowledge} {Discovery} and {Data} {Mining}}} \emph{(\bibinfo{series}{{KDD} '23})}. \bibinfo{publisher}{Association for Computing Machinery}, \bibinfo{address}{New York, NY, USA}, \bibinfo{pages}{3947--3955}.
\newblock
\showISBNx{9798400701030}
\urldef\tempurl%
\url{https://doi.org/10.1145/3580305.3599827}
\showDOI{\tempurl}


\bibitem[Edge et~al\mbox{.}(2024)]%
        {edge_local_2024}
\bibfield{author}{\bibinfo{person}{Darren Edge}, \bibinfo{person}{Ha Trinh}, \bibinfo{person}{Newman Cheng}, \bibinfo{person}{Joshua Bradley}, \bibinfo{person}{Alex Chao}, \bibinfo{person}{Apurva Mody}, \bibinfo{person}{Steven Truitt}, {and} \bibinfo{person}{Jonathan Larson}.} \bibinfo{year}{2024}\natexlab{}.
\newblock \bibinfo{title}{From {Local} to {Global}: {A} {Graph} {RAG} {Approach} to {Query}-{Focused} {Summarization}}.
\newblock
\newblock
\urldef\tempurl%
\url{https://doi.org/10.48550/arXiv.2404.16130}
\showDOI{\tempurl}
\newblock
\shownote{arXiv:2404.16130 [cs]}.


\bibitem[Fang et~al\mbox{.}(2023)]%
        {fang_recruitpro_2023}
\bibfield{author}{\bibinfo{person}{Chuyu Fang}, \bibinfo{person}{Chuan Qin}, \bibinfo{person}{Qi Zhang}, \bibinfo{person}{Kaichun Yao}, \bibinfo{person}{Jingshuai Zhang}, \bibinfo{person}{Hengshu Zhu}, \bibinfo{person}{Fuzhen Zhuang}, {and} \bibinfo{person}{Hui Xiong}.} \bibinfo{year}{2023}\natexlab{}.
\newblock \showarticletitle{{RecruitPro}: {A} {Pretrained} {Language} {Model} with {Skill}-{Aware} {Prompt} {Learning} for {Intelligent} {Recruitment}}. In \bibinfo{booktitle}{\emph{Proceedings of the 29th {ACM} {SIGKDD} {Conference} on {Knowledge} {Discovery} and {Data} {Mining}}} \emph{(\bibinfo{series}{{KDD} '23})}. \bibinfo{publisher}{Association for Computing Machinery}, \bibinfo{address}{New York, NY, USA}, \bibinfo{pages}{3991--4002}.
\newblock
\showISBNx{9798400701030}
\urldef\tempurl%
\url{https://doi.org/10.1145/3580305.3599894}
\showDOI{\tempurl}


\bibitem[Gao et~al\mbox{.}(2024)]%
        {gao_retrieval-augmented_2024}
\bibfield{author}{\bibinfo{person}{Yunfan Gao}, \bibinfo{person}{Yun Xiong}, \bibinfo{person}{Xinyu Gao}, \bibinfo{person}{Kangxiang Jia}, \bibinfo{person}{Jinliu Pan}, \bibinfo{person}{Yuxi Bi}, \bibinfo{person}{Yi Dai}, \bibinfo{person}{Jiawei Sun}, \bibinfo{person}{Qianyu Guo}, \bibinfo{person}{Meng Wang}, {and} \bibinfo{person}{Haofen Wang}.} \bibinfo{year}{2024}\natexlab{}.
\newblock \bibinfo{title}{Retrieval-{Augmented} {Generation} for {Large} {Language} {Models}: {A} {Survey}}.
\newblock
\newblock
\urldef\tempurl%
\url{https://doi.org/10.48550/arXiv.2312.10997}
\showDOI{\tempurl}
\newblock
\shownote{arXiv:2312.10997 [cs]}.


\bibitem[gkamradt(2023)]%
        {LLMTest_NeedleInAHaystack}
\bibfield{author}{\bibinfo{person}{gkamradt}.} \bibinfo{year}{2023}\natexlab{}.
\newblock \bibinfo{title}{LLMTest Needle In A Haystack - Pressure Testing LLMs}.
\newblock \bibinfo{howpublished}{\url{https://github.com/gkamradt/LLMTest_NeedleInAHaystack}}.
\newblock


\bibitem[Guo et~al\mbox{.}(2022)]%
        {guo_medical_2022}
\bibfield{author}{\bibinfo{person}{Quan Guo}, \bibinfo{person}{Shuai Cao}, {and} \bibinfo{person}{Zhang Yi}.} \bibinfo{year}{2022}\natexlab{}.
\newblock \showarticletitle{A medical question answering system using large language models and knowledge graphs}.
\newblock \bibinfo{journal}{\emph{International Journal of Intelligent Systems}} \bibinfo{volume}{37}, \bibinfo{number}{11} (\bibinfo{year}{2022}), \bibinfo{pages}{8548--8564}.
\newblock
\showISSN{1098-111X}
\urldef\tempurl%
\url{https://doi.org/10.1002/int.22955}
\showDOI{\tempurl}
\newblock
\shownote{\_eprint: https://onlinelibrary.wiley.com/doi/pdf/10.1002/int.22955}.


\bibitem[Guu et~al\mbox{.}(2020a)]%
        {guu_retrieval_2020}
\bibfield{author}{\bibinfo{person}{Kelvin Guu}, \bibinfo{person}{Kenton Lee}, \bibinfo{person}{Zora Tung}, \bibinfo{person}{Panupong Pasupat}, {and} \bibinfo{person}{Mingwei Chang}.} \bibinfo{year}{2020}\natexlab{a}.
\newblock \showarticletitle{Retrieval {Augmented} {Language} {Model} {Pre}-{Training}}. In \bibinfo{booktitle}{\emph{Proceedings of the 37th {International} {Conference} on {Machine} {Learning}}}. \bibinfo{publisher}{PMLR}, \bibinfo{pages}{3929--3938}.
\newblock
\urldef\tempurl%
\url{https://proceedings.mlr.press/v119/guu20a.html}
\showURL{%
\tempurl}
\newblock
\shownote{ISSN: 2640-3498}.


\bibitem[Guu et~al\mbox{.}(2020b)]%
        {guu_realm_2020}
\bibfield{author}{\bibinfo{person}{Kelvin Guu}, \bibinfo{person}{Kenton Lee}, \bibinfo{person}{Zora Tung}, \bibinfo{person}{Panupong Pasupat}, {and} \bibinfo{person}{Ming-Wei Chang}.} \bibinfo{year}{2020}\natexlab{b}.
\newblock \showarticletitle{{REALM}: retrieval-augmented language model pre-training}. In \bibinfo{booktitle}{\emph{Proceedings of the 37th {International} {Conference} on {Machine} {Learning}}} \emph{(\bibinfo{series}{{ICML}'20}, Vol.~\bibinfo{volume}{119})}. \bibinfo{publisher}{JMLR.org}, \bibinfo{pages}{3929--3938}.
\newblock


\bibitem[Gómez-Rodríguez and Williams(2023)]%
        {gomez-rodriguez_confederacy_2023}
\bibfield{author}{\bibinfo{person}{Carlos Gómez-Rodríguez} {and} \bibinfo{person}{Paul Williams}.} \bibinfo{year}{2023}\natexlab{}.
\newblock \showarticletitle{A {Confederacy} of {Models}: a {Comprehensive} {Evaluation} of {LLMs} on {Creative} {Writing}}. In \bibinfo{booktitle}{\emph{Findings of the {Association} for {Computational} {Linguistics}: {EMNLP} 2023}}, \bibfield{editor}{\bibinfo{person}{Houda Bouamor}, \bibinfo{person}{Juan Pino}, {and} \bibinfo{person}{Kalika Bali}} (Eds.). \bibinfo{publisher}{Association for Computational Linguistics}, \bibinfo{address}{Singapore}, \bibinfo{pages}{14504--14528}.
\newblock
\urldef\tempurl%
\url{https://doi.org/10.18653/v1/2023.findings-emnlp.966}
\showDOI{\tempurl}


\bibitem[Hamidi and Roberts(2023)]%
        {hamidi_evaluation_2023}
\bibfield{author}{\bibinfo{person}{Alaleh Hamidi} {and} \bibinfo{person}{Kirk Roberts}.} \bibinfo{year}{2023}\natexlab{}.
\newblock \bibinfo{title}{Evaluation of {AI} {Chatbots} for {Patient}-{Specific} {EHR} {Questions}}.
\newblock
\newblock
\urldef\tempurl%
\url{https://doi.org/10.48550/arXiv.2306.02549}
\showDOI{\tempurl}
\newblock
\shownote{arXiv:2306.02549 [cs]}.


\bibitem[He et~al\mbox{.}(2023)]%
        {he_survey_2023}
\bibfield{author}{\bibinfo{person}{Kai He}, \bibinfo{person}{Rui Mao}, \bibinfo{person}{Qika Lin}, \bibinfo{person}{Yucheng Ruan}, \bibinfo{person}{Xiang Lan}, \bibinfo{person}{Mengling Feng}, {and} \bibinfo{person}{Erik Cambria}.} \bibinfo{year}{2023}\natexlab{}.
\newblock \bibinfo{title}{A {Survey} of {Large} {Language} {Models} for {Healthcare}: from {Data}, {Technology}, and {Applications} to {Accountability} and {Ethics}}.
\newblock
\newblock
\urldef\tempurl%
\url{https://doi.org/10.48550/arXiv.2310.05694}
\showDOI{\tempurl}
\newblock
\shownote{arXiv:2310.05694 [cs]}.


\bibitem[Hendrycks et~al\mbox{.}(2021)]%
        {hendrycks_measuring_2021}
\bibfield{author}{\bibinfo{person}{Dan Hendrycks}, \bibinfo{person}{Collin Burns}, \bibinfo{person}{Steven Basart}, \bibinfo{person}{Andy Zou}, \bibinfo{person}{Mantas Mazeika}, \bibinfo{person}{Dawn Song}, {and} \bibinfo{person}{Jacob Steinhardt}.} \bibinfo{year}{2021}\natexlab{}.
\newblock \bibinfo{title}{Measuring {Massive} {Multitask} {Language} {Understanding}}.
\newblock
\newblock
\urldef\tempurl%
\url{https://doi.org/10.48550/arXiv.2009.03300}
\showDOI{\tempurl}
\newblock
\shownote{arXiv:2009.03300 [cs]}.


\bibitem[Houlsby et~al\mbox{.}(2019)]%
        {pmlr-v97-houlsby19a}
\bibfield{author}{\bibinfo{person}{Neil Houlsby}, \bibinfo{person}{Andrei Giurgiu}, \bibinfo{person}{Stanislaw Jastrzebski}, \bibinfo{person}{Bruna Morrone}, \bibinfo{person}{Quentin De~Laroussilhe}, \bibinfo{person}{Andrea Gesmundo}, \bibinfo{person}{Mona Attariyan}, {and} \bibinfo{person}{Sylvain Gelly}.} \bibinfo{year}{2019}\natexlab{}.
\newblock \showarticletitle{Parameter-Efficient Transfer Learning for {NLP}}. In \bibinfo{booktitle}{\emph{Proceedings of the 36th International Conference on Machine Learning}} \emph{(\bibinfo{series}{Proceedings of Machine Learning Research}, Vol.~\bibinfo{volume}{97})}, \bibfield{editor}{\bibinfo{person}{Kamalika Chaudhuri} {and} \bibinfo{person}{Ruslan Salakhutdinov}} (Eds.). \bibinfo{publisher}{PMLR}, \bibinfo{pages}{2790--2799}.
\newblock
\urldef\tempurl%
\url{https://proceedings.mlr.press/v97/houlsby19a.html}
\showURL{%
\tempurl}


\bibitem[Howard and Ruder(2018)]%
        {howard_universal_2018}
\bibfield{author}{\bibinfo{person}{Jeremy Howard} {and} \bibinfo{person}{Sebastian Ruder}.} \bibinfo{year}{2018}\natexlab{}.
\newblock \showarticletitle{Universal {Language} {Model} {Fine}-tuning for {Text} {Classification}}. In \bibinfo{booktitle}{\emph{Proceedings of the 56th {Annual} {Meeting} of the {Association} for {Computational} {Linguistics} ({Volume} 1: {Long} {Papers})}}, \bibfield{editor}{\bibinfo{person}{Iryna Gurevych} {and} \bibinfo{person}{Yusuke Miyao}} (Eds.). \bibinfo{publisher}{Association for Computational Linguistics}, \bibinfo{address}{Melbourne, Australia}, \bibinfo{pages}{328--339}.
\newblock
\urldef\tempurl%
\url{https://doi.org/10.18653/v1/P18-1031}
\showDOI{\tempurl}


\bibitem[Hu et~al\mbox{.}(2021)]%
        {hu_lora_2021}
\bibfield{author}{\bibinfo{person}{Edward~J. Hu}, \bibinfo{person}{Yelong Shen}, \bibinfo{person}{Phillip Wallis}, \bibinfo{person}{Zeyuan Allen-Zhu}, \bibinfo{person}{Yuanzhi Li}, \bibinfo{person}{Shean Wang}, \bibinfo{person}{Lu Wang}, {and} \bibinfo{person}{Weizhu Chen}.} \bibinfo{year}{2021}\natexlab{}.
\newblock \bibinfo{title}{{LoRA}: {Low}-{Rank} {Adaptation} of {Large} {Language} {Models}}.
\newblock
\newblock
\urldef\tempurl%
\url{https://doi.org/10.48550/arXiv.2106.09685}
\showDOI{\tempurl}
\newblock
\shownote{arXiv:2106.09685 [cs]}.


\bibitem[Huang et~al\mbox{.}(2023)]%
        {huang_finbert_2023}
\bibfield{author}{\bibinfo{person}{Allen~H. Huang}, \bibinfo{person}{Hui Wang}, {and} \bibinfo{person}{Yi Yang}.} \bibinfo{year}{2023}\natexlab{}.
\newblock \showarticletitle{{FinBERT}: {A} {Large} {Language} {Model} for {Extracting} {Information} from {Financial} {Text}*}.
\newblock \bibinfo{journal}{\emph{Contemporary Accounting Research}} \bibinfo{volume}{40}, \bibinfo{number}{2} (\bibinfo{year}{2023}), \bibinfo{pages}{806--841}.
\newblock
\showISSN{1911-3846}
\urldef\tempurl%
\url{https://doi.org/10.1111/1911-3846.12832}
\showDOI{\tempurl}
\newblock
\shownote{\_eprint: https://onlinelibrary.wiley.com/doi/pdf/10.1111/1911-3846.12832}.


\bibitem[Huang et~al\mbox{.}(2022)]%
        {huang_ernie-geol_2022}
\bibfield{author}{\bibinfo{person}{Jizhou Huang}, \bibinfo{person}{Haifeng Wang}, \bibinfo{person}{Yibo Sun}, \bibinfo{person}{Yunsheng Shi}, \bibinfo{person}{Zhengjie Huang}, \bibinfo{person}{An Zhuo}, {and} \bibinfo{person}{Shikun Feng}.} \bibinfo{year}{2022}\natexlab{}.
\newblock \showarticletitle{{ERNIE}-{GeoL}: {A} {Geography}-and-{Language} {Pre}-trained {Model} and its {Applications} in {Baidu} {Maps}}. In \bibinfo{booktitle}{\emph{Proceedings of the 28th {ACM} {SIGKDD} {Conference} on {Knowledge} {Discovery} and {Data} {Mining}}} \emph{(\bibinfo{series}{{KDD} '22})}. \bibinfo{publisher}{Association for Computing Machinery}, \bibinfo{address}{New York, NY, USA}, \bibinfo{pages}{3029--3039}.
\newblock
\showISBNx{978-1-4503-9385-0}
\urldef\tempurl%
\url{https://doi.org/10.1145/3534678.3539021}
\showDOI{\tempurl}


\bibitem[Kandpal et~al\mbox{.}(2023)]%
        {kandpal_large_2023}
\bibfield{author}{\bibinfo{person}{Nikhil Kandpal}, \bibinfo{person}{Haikang Deng}, \bibinfo{person}{Adam Roberts}, \bibinfo{person}{Eric Wallace}, {and} \bibinfo{person}{Colin Raffel}.} \bibinfo{year}{2023}\natexlab{}.
\newblock \showarticletitle{Large language models struggle to learn long-tail knowledge}. In \bibinfo{booktitle}{\emph{Proceedings of the 40th {International} {Conference} on {Machine} {Learning}}} \emph{(\bibinfo{series}{{ICML}'23}, Vol.~\bibinfo{volume}{202})}. \bibinfo{publisher}{JMLR.org}, \bibinfo{address}{Honolulu, Hawaii, USA}, \bibinfo{pages}{15696--15707}.
\newblock


\bibitem[Lai et~al\mbox{.}(2023)]%
        {lai_psy-llm_2023}
\bibfield{author}{\bibinfo{person}{Tin Lai}, \bibinfo{person}{Yukun Shi}, \bibinfo{person}{Zicong Du}, \bibinfo{person}{Jiajie Wu}, \bibinfo{person}{Ken Fu}, \bibinfo{person}{Yichao Dou}, {and} \bibinfo{person}{Ziqi Wang}.} \bibinfo{year}{2023}\natexlab{}.
\newblock \bibinfo{title}{Psy-{LLM}: {Scaling} up {Global} {Mental} {Health} {Psychological} {Services} with {AI}-based {Large} {Language} {Models}}.
\newblock
\newblock
\urldef\tempurl%
\url{https://doi.org/10.48550/arXiv.2307.11991}
\showDOI{\tempurl}
\newblock
\shownote{arXiv:2307.11991 [cs]}.


\bibitem[Lewis et~al\mbox{.}(2020)]%
        {lewis_retrieval-augmented_2020}
\bibfield{author}{\bibinfo{person}{Patrick Lewis}, \bibinfo{person}{Ethan Perez}, \bibinfo{person}{Aleksandra Piktus}, \bibinfo{person}{Fabio Petroni}, \bibinfo{person}{Vladimir Karpukhin}, \bibinfo{person}{Naman Goyal}, \bibinfo{person}{Heinrich Küttler}, \bibinfo{person}{Mike Lewis}, \bibinfo{person}{Wen-tau Yih}, \bibinfo{person}{Tim Rocktäschel}, \bibinfo{person}{Sebastian Riedel}, {and} \bibinfo{person}{Douwe Kiela}.} \bibinfo{year}{2020}\natexlab{}.
\newblock \showarticletitle{Retrieval-{Augmented} {Generation} for {Knowledge}-{Intensive} {NLP} {Tasks}}. In \bibinfo{booktitle}{\emph{Advances in {Neural} {Information} {Processing} {Systems}}}, Vol.~\bibinfo{volume}{33}. \bibinfo{publisher}{Curran Associates, Inc.}, \bibinfo{pages}{9459--9474}.
\newblock
\urldef\tempurl%
\url{https://proceedings.neurips.cc/paper_files/paper/2020/hash/6b493230205f780e1bc26945df7481e5-Abstract.html}
\showURL{%
\tempurl}


\bibitem[Lialin et~al\mbox{.}(2023)]%
        {lialin_scaling_2023}
\bibfield{author}{\bibinfo{person}{Vladislav Lialin}, \bibinfo{person}{Vijeta Deshpande}, {and} \bibinfo{person}{Anna Rumshisky}.} \bibinfo{year}{2023}\natexlab{}.
\newblock \bibinfo{title}{Scaling {Down} to {Scale} {Up}: {A} {Guide} to {Parameter}-{Efficient} {Fine}-{Tuning}}.
\newblock
\newblock
\urldef\tempurl%
\url{https://doi.org/10.48550/arXiv.2303.15647}
\showDOI{\tempurl}


\bibitem[Liu et~al\mbox{.}(2023)]%
        {liu_g-eval_2023}
\bibfield{author}{\bibinfo{person}{Yang Liu}, \bibinfo{person}{Dan Iter}, \bibinfo{person}{Yichong Xu}, \bibinfo{person}{Shuohang Wang}, \bibinfo{person}{Ruochen Xu}, {and} \bibinfo{person}{Chenguang Zhu}.} \bibinfo{year}{2023}\natexlab{}.
\newblock \bibinfo{title}{G-{Eval}: {NLG} {Evaluation} using {GPT}-4 with {Better} {Human} {Alignment}}.
\newblock
\newblock
\urldef\tempurl%
\url{https://doi.org/10.48550/arXiv.2303.16634}
\showDOI{\tempurl}
\newblock
\shownote{arXiv:2303.16634 [cs]}.


\bibitem[Liventsev et~al\mbox{.}(2023)]%
        {liventsev_fully_2023}
\bibfield{author}{\bibinfo{person}{Vadim Liventsev}, \bibinfo{person}{Anastasiia Grishina}, \bibinfo{person}{Aki Härmä}, {and} \bibinfo{person}{Leon Moonen}.} \bibinfo{year}{2023}\natexlab{}.
\newblock \showarticletitle{Fully {Autonomous} {Programming} with {Large} {Language} {Models}}. In \bibinfo{booktitle}{\emph{Proceedings of the {Genetic} and {Evolutionary} {Computation} {Conference}}} \emph{(\bibinfo{series}{{GECCO} '23})}. \bibinfo{publisher}{Association for Computing Machinery}, \bibinfo{address}{New York, NY, USA}, \bibinfo{pages}{1146--1155}.
\newblock
\showISBNx{9798400701191}
\urldef\tempurl%
\url{https://doi.org/10.1145/3583131.3590481}
\showDOI{\tempurl}


\bibitem[Louis et~al\mbox{.}(2023)]%
        {louis_interpretable_2023}
\bibfield{author}{\bibinfo{person}{Antoine Louis}, \bibinfo{person}{Gijs van Dijck}, {and} \bibinfo{person}{Gerasimos Spanakis}.} \bibinfo{year}{2023}\natexlab{}.
\newblock \bibinfo{title}{Interpretable {Long}-{Form} {Legal} {Question} {Answering} with {Retrieval}-{Augmented} {Large} {Language} {Models}}.
\newblock
\newblock
\urldef\tempurl%
\url{https://doi.org/10.48550/arXiv.2309.17050}
\showDOI{\tempurl}
\newblock
\shownote{arXiv:2309.17050 [cs]}.


\bibitem[Min et~al\mbox{.}(2017)]%
        {min_question_2017}
\bibfield{author}{\bibinfo{person}{Sewon Min}, \bibinfo{person}{Minjoon Seo}, {and} \bibinfo{person}{Hannaneh Hajishirzi}.} \bibinfo{year}{2017}\natexlab{}.
\newblock \showarticletitle{Question {Answering} through {Transfer} {Learning} from {Large} {Fine}-grained {Supervision} {Data}}. In \bibinfo{booktitle}{\emph{Proceedings of the 55th {Annual} {Meeting} of the {Association} for {Computational} {Linguistics} ({Volume} 2: {Short} {Papers})}}, \bibfield{editor}{\bibinfo{person}{Regina Barzilay} {and} \bibinfo{person}{Min-Yen Kan}} (Eds.). \bibinfo{publisher}{Association for Computational Linguistics}, \bibinfo{address}{Vancouver, Canada}, \bibinfo{pages}{510--517}.
\newblock
\urldef\tempurl%
\url{https://doi.org/10.18653/v1/P17-2081}
\showDOI{\tempurl}


\bibitem[Nellis and Cherney(2023)]%
        {nellis_us_2023}
\bibfield{author}{\bibinfo{person}{Stephen Nellis} {and} \bibinfo{person}{Max~A. Cherney}.} \bibinfo{year}{2023}\natexlab{}.
\newblock \showarticletitle{{US} curbs {AI} chip exports from {Nvidia} and {AMD} to some {Middle} {East} countries}.
\newblock \bibinfo{journal}{\emph{Reuters}} (\bibinfo{date}{Aug.} \bibinfo{year}{2023}).
\newblock
\urldef\tempurl%
\url{https://www.reuters.com/technology/us-restricts-exports-some-nvidia-chips-middle-east-countries-filing-2023-08-30/}
\showURL{%
\tempurl}


\bibitem[OpenAI et~al\mbox{.}(2023)]%
        {openai_gpt-4_2023}
\bibfield{author}{\bibinfo{person}{OpenAI}, \bibinfo{person}{Josh Achiam}, \bibinfo{person}{Steven Adler}, \bibinfo{person}{Sandhini Agarwal}, \bibinfo{person}{Lama Ahmad}, \bibinfo{person}{Ilge Akkaya}, \bibinfo{person}{Florencia~Leoni Aleman}, \bibinfo{person}{Diogo Almeida}, \bibinfo{person}{Janko Altenschmidt}, \bibinfo{person}{Sam Altman}, \bibinfo{person}{Shyamal Anadkat}, \bibinfo{person}{Red Avila}, \bibinfo{person}{Igor Babuschkin}, \bibinfo{person}{Suchir Balaji}, \bibinfo{person}{Valerie Balcom}, \bibinfo{person}{Paul Baltescu}, \bibinfo{person}{Haiming Bao}, \bibinfo{person}{Mo Bavarian}, \bibinfo{person}{Jeff Belgum}, \bibinfo{person}{Irwan Bello}, \bibinfo{person}{Jake Berdine}, \bibinfo{person}{Gabriel Bernadett-Shapiro}, \bibinfo{person}{Christopher Berner}, \bibinfo{person}{Lenny Bogdonoff}, \bibinfo{person}{Oleg Boiko}, \bibinfo{person}{Madelaine Boyd}, \bibinfo{person}{Anna-Luisa Brakman}, \bibinfo{person}{Greg Brockman}, \bibinfo{person}{Tim Brooks}, \bibinfo{person}{Miles Brundage},
  \bibinfo{person}{Kevin Button}, \bibinfo{person}{Trevor Cai}, \bibinfo{person}{Rosie Campbell}, \bibinfo{person}{Andrew Cann}, \bibinfo{person}{Brittany Carey}, \bibinfo{person}{Chelsea Carlson}, \bibinfo{person}{Rory Carmichael}, \bibinfo{person}{Brooke Chan}, \bibinfo{person}{Che Chang}, \bibinfo{person}{Fotis Chantzis}, \bibinfo{person}{Derek Chen}, \bibinfo{person}{Sully Chen}, \bibinfo{person}{Ruby Chen}, \bibinfo{person}{Jason Chen}, \bibinfo{person}{Mark Chen}, \bibinfo{person}{Ben Chess}, \bibinfo{person}{Chester Cho}, \bibinfo{person}{Casey Chu}, \bibinfo{person}{Hyung~Won Chung}, \bibinfo{person}{Dave Cummings}, \bibinfo{person}{Jeremiah Currier}, \bibinfo{person}{Yunxing Dai}, \bibinfo{person}{Cory Decareaux}, \bibinfo{person}{Thomas Degry}, \bibinfo{person}{Noah Deutsch}, \bibinfo{person}{Damien Deville}, \bibinfo{person}{Arka Dhar}, \bibinfo{person}{David Dohan}, \bibinfo{person}{Steve Dowling}, \bibinfo{person}{Sheila Dunning}, \bibinfo{person}{Adrien Ecoffet}, \bibinfo{person}{Atty Eleti},
  \bibinfo{person}{Tyna Eloundou}, \bibinfo{person}{David Farhi}, \bibinfo{person}{Liam Fedus}, \bibinfo{person}{Niko Felix}, \bibinfo{person}{Simón~Posada Fishman}, \bibinfo{person}{Juston Forte}, \bibinfo{person}{Isabella Fulford}, \bibinfo{person}{Leo Gao}, \bibinfo{person}{Elie Georges}, \bibinfo{person}{Christian Gibson}, \bibinfo{person}{Vik Goel}, \bibinfo{person}{Tarun Gogineni}, \bibinfo{person}{Gabriel Goh}, \bibinfo{person}{Rapha Gontijo-Lopes}, \bibinfo{person}{Jonathan Gordon}, \bibinfo{person}{Morgan Grafstein}, \bibinfo{person}{Scott Gray}, \bibinfo{person}{Ryan Greene}, \bibinfo{person}{Joshua Gross}, \bibinfo{person}{Shixiang~Shane Gu}, \bibinfo{person}{Yufei Guo}, \bibinfo{person}{Chris Hallacy}, \bibinfo{person}{Jesse Han}, \bibinfo{person}{Jeff Harris}, \bibinfo{person}{Yuchen He}, \bibinfo{person}{Mike Heaton}, \bibinfo{person}{Johannes Heidecke}, \bibinfo{person}{Chris Hesse}, \bibinfo{person}{Alan Hickey}, \bibinfo{person}{Wade Hickey}, \bibinfo{person}{Peter Hoeschele},
  \bibinfo{person}{Brandon Houghton}, \bibinfo{person}{Kenny Hsu}, \bibinfo{person}{Shengli Hu}, \bibinfo{person}{Xin Hu}, \bibinfo{person}{Joost Huizinga}, \bibinfo{person}{Shantanu Jain}, \bibinfo{person}{Shawn Jain}, \bibinfo{person}{Joanne Jang}, \bibinfo{person}{Angela Jiang}, \bibinfo{person}{Roger Jiang}, \bibinfo{person}{Haozhun Jin}, \bibinfo{person}{Denny Jin}, \bibinfo{person}{Shino Jomoto}, \bibinfo{person}{Billie Jonn}, \bibinfo{person}{Heewoo Jun}, \bibinfo{person}{Tomer Kaftan}, \bibinfo{person}{Łukasz Kaiser}, \bibinfo{person}{Ali Kamali}, \bibinfo{person}{Ingmar Kanitscheider}, \bibinfo{person}{Nitish~Shirish Keskar}, \bibinfo{person}{Tabarak Khan}, \bibinfo{person}{Logan Kilpatrick}, \bibinfo{person}{Jong~Wook Kim}, \bibinfo{person}{Christina Kim}, \bibinfo{person}{Yongjik Kim}, \bibinfo{person}{Hendrik Kirchner}, \bibinfo{person}{Jamie Kiros}, \bibinfo{person}{Matt Knight}, \bibinfo{person}{Daniel Kokotajlo}, \bibinfo{person}{Łukasz Kondraciuk}, \bibinfo{person}{Andrew Kondrich},
  \bibinfo{person}{Aris Konstantinidis}, \bibinfo{person}{Kyle Kosic}, \bibinfo{person}{Gretchen Krueger}, \bibinfo{person}{Vishal Kuo}, \bibinfo{person}{Michael Lampe}, \bibinfo{person}{Ikai Lan}, \bibinfo{person}{Teddy Lee}, \bibinfo{person}{Jan Leike}, \bibinfo{person}{Jade Leung}, \bibinfo{person}{Daniel Levy}, \bibinfo{person}{Chak~Ming Li}, \bibinfo{person}{Rachel Lim}, \bibinfo{person}{Molly Lin}, \bibinfo{person}{Stephanie Lin}, \bibinfo{person}{Mateusz Litwin}, \bibinfo{person}{Theresa Lopez}, \bibinfo{person}{Ryan Lowe}, \bibinfo{person}{Patricia Lue}, \bibinfo{person}{Anna Makanju}, \bibinfo{person}{Kim Malfacini}, \bibinfo{person}{Sam Manning}, \bibinfo{person}{Todor Markov}, \bibinfo{person}{Yaniv Markovski}, \bibinfo{person}{Bianca Martin}, \bibinfo{person}{Katie Mayer}, \bibinfo{person}{Andrew Mayne}, \bibinfo{person}{Bob McGrew}, \bibinfo{person}{Scott~Mayer McKinney}, \bibinfo{person}{Christine McLeavey}, \bibinfo{person}{Paul McMillan}, \bibinfo{person}{Jake McNeil}, \bibinfo{person}{David
  Medina}, \bibinfo{person}{Aalok Mehta}, \bibinfo{person}{Jacob Menick}, \bibinfo{person}{Luke Metz}, \bibinfo{person}{Andrey Mishchenko}, \bibinfo{person}{Pamela Mishkin}, \bibinfo{person}{Vinnie Monaco}, \bibinfo{person}{Evan Morikawa}, \bibinfo{person}{Daniel Mossing}, \bibinfo{person}{Tong Mu}, \bibinfo{person}{Mira Murati}, \bibinfo{person}{Oleg Murk}, \bibinfo{person}{David Mély}, \bibinfo{person}{Ashvin Nair}, \bibinfo{person}{Reiichiro Nakano}, \bibinfo{person}{Rajeev Nayak}, \bibinfo{person}{Arvind Neelakantan}, \bibinfo{person}{Richard Ngo}, \bibinfo{person}{Hyeonwoo Noh}, \bibinfo{person}{Long Ouyang}, \bibinfo{person}{Cullen O'Keefe}, \bibinfo{person}{Jakub Pachocki}, \bibinfo{person}{Alex Paino}, \bibinfo{person}{Joe Palermo}, \bibinfo{person}{Ashley Pantuliano}, \bibinfo{person}{Giambattista Parascandolo}, \bibinfo{person}{Joel Parish}, \bibinfo{person}{Emy Parparita}, \bibinfo{person}{Alex Passos}, \bibinfo{person}{Mikhail Pavlov}, \bibinfo{person}{Andrew Peng}, \bibinfo{person}{Adam
  Perelman}, \bibinfo{person}{Filipe de Avila~Belbute Peres}, \bibinfo{person}{Michael Petrov}, \bibinfo{person}{Henrique Ponde de~Oliveira Pinto}, \bibinfo{person}{Michael}, \bibinfo{person}{Pokorny}, \bibinfo{person}{Michelle Pokrass}, \bibinfo{person}{Vitchyr Pong}, \bibinfo{person}{Tolly Powell}, \bibinfo{person}{Alethea Power}, \bibinfo{person}{Boris Power}, \bibinfo{person}{Elizabeth Proehl}, \bibinfo{person}{Raul Puri}, \bibinfo{person}{Alec Radford}, \bibinfo{person}{Jack Rae}, \bibinfo{person}{Aditya Ramesh}, \bibinfo{person}{Cameron Raymond}, \bibinfo{person}{Francis Real}, \bibinfo{person}{Kendra Rimbach}, \bibinfo{person}{Carl Ross}, \bibinfo{person}{Bob Rotsted}, \bibinfo{person}{Henri Roussez}, \bibinfo{person}{Nick Ryder}, \bibinfo{person}{Mario Saltarelli}, \bibinfo{person}{Ted Sanders}, \bibinfo{person}{Shibani Santurkar}, \bibinfo{person}{Girish Sastry}, \bibinfo{person}{Heather Schmidt}, \bibinfo{person}{David Schnurr}, \bibinfo{person}{John Schulman}, \bibinfo{person}{Daniel Selsam},
  \bibinfo{person}{Kyla Sheppard}, \bibinfo{person}{Toki Sherbakov}, \bibinfo{person}{Jessica Shieh}, \bibinfo{person}{Sarah Shoker}, \bibinfo{person}{Pranav Shyam}, \bibinfo{person}{Szymon Sidor}, \bibinfo{person}{Eric Sigler}, \bibinfo{person}{Maddie Simens}, \bibinfo{person}{Jordan Sitkin}, \bibinfo{person}{Katarina Slama}, \bibinfo{person}{Ian Sohl}, \bibinfo{person}{Benjamin Sokolowsky}, \bibinfo{person}{Yang Song}, \bibinfo{person}{Natalie Staudacher}, \bibinfo{person}{Felipe~Petroski Such}, \bibinfo{person}{Natalie Summers}, \bibinfo{person}{Ilya Sutskever}, \bibinfo{person}{Jie Tang}, \bibinfo{person}{Nikolas Tezak}, \bibinfo{person}{Madeleine Thompson}, \bibinfo{person}{Phil Tillet}, \bibinfo{person}{Amin Tootoonchian}, \bibinfo{person}{Elizabeth Tseng}, \bibinfo{person}{Preston Tuggle}, \bibinfo{person}{Nick Turley}, \bibinfo{person}{Jerry Tworek}, \bibinfo{person}{Juan Felipe~Cerón Uribe}, \bibinfo{person}{Andrea Vallone}, \bibinfo{person}{Arun Vijayvergiya}, \bibinfo{person}{Chelsea Voss},
  \bibinfo{person}{Carroll Wainwright}, \bibinfo{person}{Justin~Jay Wang}, \bibinfo{person}{Alvin Wang}, \bibinfo{person}{Ben Wang}, \bibinfo{person}{Jonathan Ward}, \bibinfo{person}{Jason Wei}, \bibinfo{person}{C.~J. Weinmann}, \bibinfo{person}{Akila Welihinda}, \bibinfo{person}{Peter Welinder}, \bibinfo{person}{Jiayi Weng}, \bibinfo{person}{Lilian Weng}, \bibinfo{person}{Matt Wiethoff}, \bibinfo{person}{Dave Willner}, \bibinfo{person}{Clemens Winter}, \bibinfo{person}{Samuel Wolrich}, \bibinfo{person}{Hannah Wong}, \bibinfo{person}{Lauren Workman}, \bibinfo{person}{Sherwin Wu}, \bibinfo{person}{Jeff Wu}, \bibinfo{person}{Michael Wu}, \bibinfo{person}{Kai Xiao}, \bibinfo{person}{Tao Xu}, \bibinfo{person}{Sarah Yoo}, \bibinfo{person}{Kevin Yu}, \bibinfo{person}{Qiming Yuan}, \bibinfo{person}{Wojciech Zaremba}, \bibinfo{person}{Rowan Zellers}, \bibinfo{person}{Chong Zhang}, \bibinfo{person}{Marvin Zhang}, \bibinfo{person}{Shengjia Zhao}, \bibinfo{person}{Tianhao Zheng}, \bibinfo{person}{Juntang Zhuang},
  \bibinfo{person}{William Zhuk}, {and} \bibinfo{person}{Barret Zoph}.} \bibinfo{year}{2023}\natexlab{}.
\newblock \bibinfo{title}{{GPT}-4 {Technical} {Report}}.
\newblock
\newblock
\urldef\tempurl%
\url{https://doi.org/10.48550/arXiv.2303.08774}
\showDOI{\tempurl}
\newblock
\shownote{arXiv:2303.08774 [cs]}.


\bibitem[Ram et~al\mbox{.}(2023)]%
        {ram_-context_2023}
\bibfield{author}{\bibinfo{person}{Ori Ram}, \bibinfo{person}{Yoav Levine}, \bibinfo{person}{Itay Dalmedigos}, \bibinfo{person}{Dor Muhlgay}, \bibinfo{person}{Amnon Shashua}, \bibinfo{person}{Kevin Leyton-Brown}, {and} \bibinfo{person}{Yoav Shoham}.} \bibinfo{year}{2023}\natexlab{}.
\newblock \bibinfo{title}{In-{Context} {Retrieval}-{Augmented} {Language} {Models}}.
\newblock
\newblock
\urldef\tempurl%
\url{https://doi.org/10.48550/arXiv.2302.00083}
\showDOI{\tempurl}
\newblock
\shownote{arXiv:2302.00083 [cs]}.


\bibitem[Su et~al\mbox{.}(2023)]%
        {su-etal-2023-one}
\bibfield{author}{\bibinfo{person}{Hongjin Su}, \bibinfo{person}{Weijia Shi}, \bibinfo{person}{Jungo Kasai}, \bibinfo{person}{Yizhong Wang}, \bibinfo{person}{Yushi Hu}, \bibinfo{person}{Mari Ostendorf}, \bibinfo{person}{Wen-tau Yih}, \bibinfo{person}{Noah~A. Smith}, \bibinfo{person}{Luke Zettlemoyer}, {and} \bibinfo{person}{Tao Yu}.} \bibinfo{year}{2023}\natexlab{}.
\newblock \showarticletitle{One Embedder, Any Task: Instruction-Finetuned Text Embeddings}. In \bibinfo{booktitle}{\emph{Findings of the Association for Computational Linguistics: ACL 2023}}, \bibfield{editor}{\bibinfo{person}{Anna Rogers}, \bibinfo{person}{Jordan Boyd-Graber}, {and} \bibinfo{person}{Naoaki Okazaki}} (Eds.). \bibinfo{publisher}{Association for Computational Linguistics}, \bibinfo{address}{Toronto, Canada}, \bibinfo{pages}{1102--1121}.
\newblock
\urldef\tempurl%
\url{https://doi.org/10.18653/v1/2023.findings-acl.71}
\showDOI{\tempurl}


\bibitem[Touvron et~al\mbox{.}(2023)]%
        {touvron_llama_2023}
\bibfield{author}{\bibinfo{person}{Hugo Touvron}, \bibinfo{person}{Louis Martin}, \bibinfo{person}{Kevin Stone}, \bibinfo{person}{Peter Albert}, \bibinfo{person}{Amjad Almahairi}, \bibinfo{person}{Yasmine Babaei}, \bibinfo{person}{Nikolay Bashlykov}, \bibinfo{person}{Soumya Batra}, \bibinfo{person}{Prajjwal Bhargava}, \bibinfo{person}{Shruti Bhosale}, \bibinfo{person}{Dan Bikel}, \bibinfo{person}{Lukas Blecher}, \bibinfo{person}{Cristian~Canton Ferrer}, \bibinfo{person}{Moya Chen}, \bibinfo{person}{Guillem Cucurull}, \bibinfo{person}{David Esiobu}, \bibinfo{person}{Jude Fernandes}, \bibinfo{person}{Jeremy Fu}, \bibinfo{person}{Wenyin Fu}, \bibinfo{person}{Brian Fuller}, \bibinfo{person}{Cynthia Gao}, \bibinfo{person}{Vedanuj Goswami}, \bibinfo{person}{Naman Goyal}, \bibinfo{person}{Anthony Hartshorn}, \bibinfo{person}{Saghar Hosseini}, \bibinfo{person}{Rui Hou}, \bibinfo{person}{Hakan Inan}, \bibinfo{person}{Marcin Kardas}, \bibinfo{person}{Viktor Kerkez}, \bibinfo{person}{Madian Khabsa},
  \bibinfo{person}{Isabel Kloumann}, \bibinfo{person}{Artem Korenev}, \bibinfo{person}{Punit~Singh Koura}, \bibinfo{person}{Marie-Anne Lachaux}, \bibinfo{person}{Thibaut Lavril}, \bibinfo{person}{Jenya Lee}, \bibinfo{person}{Diana Liskovich}, \bibinfo{person}{Yinghai Lu}, \bibinfo{person}{Yuning Mao}, \bibinfo{person}{Xavier Martinet}, \bibinfo{person}{Todor Mihaylov}, \bibinfo{person}{Pushkar Mishra}, \bibinfo{person}{Igor Molybog}, \bibinfo{person}{Yixin Nie}, \bibinfo{person}{Andrew Poulton}, \bibinfo{person}{Jeremy Reizenstein}, \bibinfo{person}{Rashi Rungta}, \bibinfo{person}{Kalyan Saladi}, \bibinfo{person}{Alan Schelten}, \bibinfo{person}{Ruan Silva}, \bibinfo{person}{Eric~Michael Smith}, \bibinfo{person}{Ranjan Subramanian}, \bibinfo{person}{Xiaoqing~Ellen Tan}, \bibinfo{person}{Binh Tang}, \bibinfo{person}{Ross Taylor}, \bibinfo{person}{Adina Williams}, \bibinfo{person}{Jian~Xiang Kuan}, \bibinfo{person}{Puxin Xu}, \bibinfo{person}{Zheng Yan}, \bibinfo{person}{Iliyan Zarov}, \bibinfo{person}{Yuchen
  Zhang}, \bibinfo{person}{Angela Fan}, \bibinfo{person}{Melanie Kambadur}, \bibinfo{person}{Sharan Narang}, \bibinfo{person}{Aurelien Rodriguez}, \bibinfo{person}{Robert Stojnic}, \bibinfo{person}{Sergey Edunov}, {and} \bibinfo{person}{Thomas Scialom}.} \bibinfo{year}{2023}\natexlab{}.
\newblock \bibinfo{title}{Llama 2: {Open} {Foundation} and {Fine}-{Tuned} {Chat} {Models}}.
\newblock
\newblock
\urldef\tempurl%
\url{https://doi.org/10.48550/arXiv.2307.09288}
\showDOI{\tempurl}
\newblock
\shownote{arXiv:2307.09288 [cs]}.


\bibitem[Vaswani et~al\mbox{.}(2017)]%
        {vaswani_attention_2017}
\bibfield{author}{\bibinfo{person}{Ashish Vaswani}, \bibinfo{person}{Noam Shazeer}, \bibinfo{person}{Niki Parmar}, \bibinfo{person}{Jakob Uszkoreit}, \bibinfo{person}{Llion Jones}, \bibinfo{person}{Aidan~N Gomez}, \bibinfo{person}{Łukasz Kaiser}, {and} \bibinfo{person}{Illia Polosukhin}.} \bibinfo{year}{2017}\natexlab{}.
\newblock \showarticletitle{Attention is {All} you {Need}}. In \bibinfo{booktitle}{\emph{Advances in {Neural} {Information} {Processing} {Systems}}}, Vol.~\bibinfo{volume}{30}. \bibinfo{publisher}{Curran Associates, Inc.}
\newblock
\urldef\tempurl%
\url{https://proceedings.neurips.cc/paper_files/paper/2017/hash/3f5ee243547dee91fbd053c1c4a845aa-Abstract.html}
\showURL{%
\tempurl}


\bibitem[Wang et~al\mbox{.}(2023)]%
        {wang_pandalm_2023}
\bibfield{author}{\bibinfo{person}{Yidong Wang}, \bibinfo{person}{Zhuohao Yu}, \bibinfo{person}{Zhengran Zeng}, \bibinfo{person}{Linyi Yang}, \bibinfo{person}{Cunxiang Wang}, \bibinfo{person}{Hao Chen}, \bibinfo{person}{Chaoya Jiang}, \bibinfo{person}{Rui Xie}, \bibinfo{person}{Jindong Wang}, \bibinfo{person}{Xing Xie}, \bibinfo{person}{Wei Ye}, \bibinfo{person}{Shikun Zhang}, {and} \bibinfo{person}{Yue Zhang}.} \bibinfo{year}{2023}\natexlab{}.
\newblock \bibinfo{title}{{PandaLM}: {An} {Automatic} {Evaluation} {Benchmark} for {LLM} {Instruction} {Tuning} {Optimization}}.
\newblock
\newblock
\urldef\tempurl%
\url{https://doi.org/10.48550/arXiv.2306.05087}
\showDOI{\tempurl}
\newblock
\shownote{arXiv:2306.05087 [cs]}.


\bibitem[Wu et~al\mbox{.}(2023)]%
        {wu_retrieve-rewrite-answer_2023}
\bibfield{author}{\bibinfo{person}{Yike Wu}, \bibinfo{person}{Nan Hu}, \bibinfo{person}{Sheng Bi}, \bibinfo{person}{Guilin Qi}, \bibinfo{person}{Jie Ren}, \bibinfo{person}{Anhuan Xie}, {and} \bibinfo{person}{Wei Song}.} \bibinfo{year}{2023}\natexlab{}.
\newblock \bibinfo{title}{Retrieve-{Rewrite}-{Answer}: {A} {KG}-to-{Text} {Enhanced} {LLMs} {Framework} for {Knowledge} {Graph} {Question} {Answering}}.
\newblock
\newblock
\urldef\tempurl%
\url{https://doi.org/10.48550/arXiv.2309.11206}
\showDOI{\tempurl}
\newblock
\shownote{arXiv:2309.11206 [cs]}.


\bibitem[Xia et~al\mbox{.}(2022)]%
        {xia_medconqa_2022}
\bibfield{author}{\bibinfo{person}{Fei Xia}, \bibinfo{person}{Bin Li}, \bibinfo{person}{Yixuan Weng}, \bibinfo{person}{Shizhu He}, \bibinfo{person}{Kang Liu}, \bibinfo{person}{Bin Sun}, \bibinfo{person}{Shutao Li}, {and} \bibinfo{person}{Jun Zhao}.} \bibinfo{year}{2022}\natexlab{}.
\newblock \showarticletitle{{MedConQA}: {Medical} {Conversational} {Question} {Answering} {System} based on {Knowledge} {Graphs}}. In \bibinfo{booktitle}{\emph{Proceedings of the 2022 {Conference} on {Empirical} {Methods} in {Natural} {Language} {Processing}: {System} {Demonstrations}}}, \bibfield{editor}{\bibinfo{person}{Wanxiang Che} {and} \bibinfo{person}{Ekaterina Shutova}} (Eds.). \bibinfo{publisher}{Association for Computational Linguistics}, \bibinfo{address}{Abu Dhabi, UAE}, \bibinfo{pages}{148--158}.
\newblock
\urldef\tempurl%
\url{https://doi.org/10.18653/v1/2022.emnlp-demos.15}
\showDOI{\tempurl}


\bibitem[Xiao et~al\mbox{.}(2022)]%
        {xiao_training_2022}
\bibfield{author}{\bibinfo{person}{Shitao Xiao}, \bibinfo{person}{Zheng Liu}, \bibinfo{person}{Yingxia Shao}, \bibinfo{person}{Tao Di}, \bibinfo{person}{Bhuvan Middha}, \bibinfo{person}{Fangzhao Wu}, {and} \bibinfo{person}{Xing Xie}.} \bibinfo{year}{2022}\natexlab{}.
\newblock \showarticletitle{Training {Large}-{Scale} {News} {Recommenders} with {Pretrained} {Language} {Models} in the {Loop}}. In \bibinfo{booktitle}{\emph{Proceedings of the 28th {ACM} {SIGKDD} {Conference} on {Knowledge} {Discovery} and {Data} {Mining}}} \emph{(\bibinfo{series}{{KDD} '22})}. \bibinfo{publisher}{Association for Computing Machinery}, \bibinfo{address}{New York, NY, USA}, \bibinfo{pages}{4215--4225}.
\newblock
\showISBNx{978-1-4503-9385-0}
\urldef\tempurl%
\url{https://doi.org/10.1145/3534678.3539120}
\showDOI{\tempurl}


\bibitem[Xu et~al\mbox{.}(2023)]%
        {xu_parameter-efficient_2023}
\bibfield{author}{\bibinfo{person}{Lingling Xu}, \bibinfo{person}{Haoran Xie}, \bibinfo{person}{Si-Zhao~Joe Qin}, \bibinfo{person}{Xiaohui Tao}, {and} \bibinfo{person}{Fu~Lee Wang}.} \bibinfo{year}{2023}\natexlab{}.
\newblock \bibinfo{title}{Parameter-{Efficient} {Fine}-{Tuning} {Methods} for {Pretrained} {Language} {Models}: {A} {Critical} {Review} and {Assessment}}.
\newblock
\newblock
\urldef\tempurl%
\url{https://doi.org/10.48550/arXiv.2312.12148}
\showDOI{\tempurl}
\newblock
\shownote{arXiv:2312.12148 [cs]}.


\bibitem[Xu et~al\mbox{.}(2024)]%
        {xu_retrieval_2024}
\bibfield{author}{\bibinfo{person}{Peng Xu}, \bibinfo{person}{Wei Ping}, \bibinfo{person}{Xianchao Wu}, \bibinfo{person}{Lawrence McAfee}, \bibinfo{person}{Chen Zhu}, \bibinfo{person}{Zihan Liu}, \bibinfo{person}{Sandeep Subramanian}, \bibinfo{person}{Evelina Bakhturina}, \bibinfo{person}{Mohammad Shoeybi}, {and} \bibinfo{person}{Bryan Catanzaro}.} \bibinfo{year}{2024}\natexlab{}.
\newblock \bibinfo{title}{Retrieval meets {Long} {Context} {Large} {Language} {Models}}.
\newblock
\newblock
\urldef\tempurl%
\url{https://doi.org/10.48550/arXiv.2310.03025}
\showDOI{\tempurl}
\newblock
\shownote{arXiv:2310.03025 [cs]}.


\bibitem[Yang et~al\mbox{.}(2023)]%
        {yang_empower_2023}
\bibfield{author}{\bibinfo{person}{Fangkai Yang}, \bibinfo{person}{Pu Zhao}, \bibinfo{person}{Zezhong Wang}, \bibinfo{person}{Lu Wang}, \bibinfo{person}{Jue Zhang}, \bibinfo{person}{Mohit Garg}, \bibinfo{person}{Qingwei Lin}, \bibinfo{person}{Saravan Rajmohan}, {and} \bibinfo{person}{Dongmei Zhang}.} \bibinfo{year}{2023}\natexlab{}.
\newblock \bibinfo{title}{Empower {Large} {Language} {Model} to {Perform} {Better} on {Industrial} {Domain}-{Specific} {Question} {Answering}}.
\newblock
\newblock
\urldef\tempurl%
\url{http://arxiv.org/abs/2305.11541}
\showURL{%
\tempurl}
\newblock
\shownote{arXiv:2305.11541 [cs]}.


\bibitem[Ye et~al\mbox{.}(2023)]%
        {ye_cognitive_2023}
\bibfield{author}{\bibinfo{person}{Hongbin Ye}, \bibinfo{person}{Tong Liu}, \bibinfo{person}{Aijia Zhang}, \bibinfo{person}{Wei Hua}, {and} \bibinfo{person}{Weiqiang Jia}.} \bibinfo{year}{2023}\natexlab{}.
\newblock \bibinfo{title}{Cognitive {Mirage}: {A} {Review} of {Hallucinations} in {Large} {Language} {Models}}.
\newblock
\newblock
\urldef\tempurl%
\url{https://doi.org/10.48550/arXiv.2309.06794}
\showDOI{\tempurl}
\newblock
\shownote{arXiv:2309.06794 [cs]}.


\bibitem[Zhang et~al\mbox{.}(2024)]%
        {zhang_extending_2024}
\bibfield{author}{\bibinfo{person}{Yikai Zhang}, \bibinfo{person}{Junlong Li}, {and} \bibinfo{person}{Pengfei Liu}.} \bibinfo{year}{2024}\natexlab{}.
\newblock \bibinfo{title}{Extending {LLMs}' {Context} {Window} with 100 {Samples}}.
\newblock
\newblock
\urldef\tempurl%
\url{https://doi.org/10.48550/arXiv.2401.07004}
\showDOI{\tempurl}
\newblock
\shownote{arXiv:2401.07004 [cs]}.


\bibitem[Zhang et~al\mbox{.}(2023)]%
        {zhang_sirens_2023}
\bibfield{author}{\bibinfo{person}{Yue Zhang}, \bibinfo{person}{Yafu Li}, \bibinfo{person}{Leyang Cui}, \bibinfo{person}{Deng Cai}, \bibinfo{person}{Lemao Liu}, \bibinfo{person}{Tingchen Fu}, \bibinfo{person}{Xinting Huang}, \bibinfo{person}{Enbo Zhao}, \bibinfo{person}{Yu Zhang}, \bibinfo{person}{Yulong Chen}, \bibinfo{person}{Longyue Wang}, \bibinfo{person}{Anh~Tuan Luu}, \bibinfo{person}{Wei Bi}, \bibinfo{person}{Freda Shi}, {and} \bibinfo{person}{Shuming Shi}.} \bibinfo{year}{2023}\natexlab{}.
\newblock \bibinfo{title}{Siren's {Song} in the {AI} {Ocean}: {A} {Survey} on {Hallucination} in {Large} {Language} {Models}}.
\newblock
\newblock
\urldef\tempurl%
\url{https://doi.org/10.48550/arXiv.2309.01219}
\showDOI{\tempurl}
\newblock
\shownote{arXiv:2309.01219 [cs]}.


\bibitem[Zhao et~al\mbox{.}(2023)]%
        {zhao_survey_2023}
\bibfield{author}{\bibinfo{person}{Wayne~Xin Zhao}, \bibinfo{person}{Kun Zhou}, \bibinfo{person}{Junyi Li}, \bibinfo{person}{Tianyi Tang}, \bibinfo{person}{Xiaolei Wang}, \bibinfo{person}{Yupeng Hou}, \bibinfo{person}{Yingqian Min}, \bibinfo{person}{Beichen Zhang}, \bibinfo{person}{Junjie Zhang}, \bibinfo{person}{Zican Dong}, \bibinfo{person}{Yifan Du}, \bibinfo{person}{Chen Yang}, \bibinfo{person}{Yushuo Chen}, \bibinfo{person}{Zhipeng Chen}, \bibinfo{person}{Jinhao Jiang}, \bibinfo{person}{Ruiyang Ren}, \bibinfo{person}{Yifan Li}, \bibinfo{person}{Xinyu Tang}, \bibinfo{person}{Zikang Liu}, \bibinfo{person}{Peiyu Liu}, \bibinfo{person}{Jian-Yun Nie}, {and} \bibinfo{person}{Ji-Rong Wen}.} \bibinfo{year}{2023}\natexlab{}.
\newblock \bibinfo{title}{A {Survey} of {Large} {Language} {Models}}.
\newblock
\newblock
\urldef\tempurl%
\url{https://doi.org/10.48550/arXiv.2303.18223}
\showDOI{\tempurl}
\newblock
\shownote{arXiv:2303.18223 [cs]}.


\end{thebibliography}

\clearpage
\appendix

\section{Prompts}

\subsection{Prompts for Question and Answer Instruction Dataset Generation}
\label{sec:instruction_dataset_prompts}

\begin{oframed}
\textbf{PROMPT1:}

As an AI assistant, your task is To generate question types, questions, answers, and source text for answers that comprehensively cover a given text and present them in a table format, considering various question types such as dialogs, conversations, conversations summary, conversations fact check, fact check, True or False, summary, quiz, review, Fill in the Blank, Short Answer, Yes or No, Matrix, compound questions, etc., follow these steps:

\begin{enumerate}
    \item Read the given text: Carefully read the text to understand its content and context.
    \item Identify key information: Look for definitions, metrics, facts, figures, events, or statements that can be used to create questions, ensuring that all parts of the text are covered.
    \item Determine question types: Based on the key information, decide which question types would be most suitable for the text (dialogs, conversations, conversations summary, conversations fact check, fact check, True or False, summary, quiz, review, Fill in the Blank, Short Answer, Yes or No, Matrix, compound questions, classify question, etc.).
    \item Generate questions: Create questions according to the chosen question types, ensuring they are clear, concise, and aligned with the question type. Make sure the questions cover the entire text.
    \item Locate answers: Find the corresponding answers for the questions within the text.
    \item Obtain source text: Replicate the essential text sections that offer the solutions to the queries, functioning as the basis for the responses. The source text must be thoroughly exhibited in a complete manner. It is necessary for the source text to be presented in a clear and understandable English without utilizing ellipsis.
    \item Design a table: Create a table with columns for question type, question, answer, and source text.
    \item Populate the table: Fill the table with the generated question types, questions, answers, and the extracted source text.
    \item Review and revise: Double-check the table for accuracy, comprehensiveness, and make any necessary adjustments.
    
\end{enumerate}

Here's an example of how to structure your table:

| Question Type | Question | Answer | Source Text |

|------------------------|----------|--------|----------------------|

| Dialog | Q1 | A1 | Source Text 1 |

| Conversation | Q2 | A2 | Source Text 2 |

| Conversation Summary | Q3 | A3 | Source Text 3 |

| Conversation Fact Check| Q4 | A4 | Source Text 4 |

| Fact Check | Q5 | A5 | Source Text 5 |

| True or False | Q6 | A6 | Source Text 6 |

| Summary | Q7 | A7 | Source Text 7 |

| Quiz | Q8 | A8 | Source Text 8 |

| Review | Q9 | A9 | Source Text 9 |

| Fill in the Blank | Q10 | A10 | Source Text 10 |

| Short Answer | Q11 | A11 | Source Text 11 |

| Yes or No | Q12 | A12 | Source Text 12 |

| Matrix | Q13 | A13 | Source Text 13 |

| Compound Question | Q14 | A14 | Source Text 14 |

| ... | ... | ... | ... |

\end{oframed}

\begin{oframed}
\textbf{PROMPT2:}

As an AI assistant, your task is To generate a Q\&A conversation between a user and an assistant using a given text, follow these steps:

\begin{enumerate}
    \item Read and understand the text: Carefully read the given text and make sure you understand the key points, facts, and ideas presented in it.
    \item Identify potential questions: Based on your understanding of the text, think of possible questions that a user might ask related to the content. Focus on questions that cover the main ideas and details from the text. Try to come up with at least 2-3 questions per topic.
    \item Formulate clear and concise answers: For each question, draft a clear and concise answer based on the information provided in the text. Make sure your answers are accurate, informative, and directly address the questions.
    \item Organize the conversation: Arrange the questions and answers into a logical order, creating a smooth flow for the conversation. Use a back-and-forth format where the user asks a question and the AI assistant provides an answer, followed by another question and answer, and so on. Ensure that there are at least 2-3 exchanges per topic.
    \item Check for clarity and coherence: Review the conversation to ensure that it is clear, coherent, and easy to follow. Make sure that the questions and answers are related to the given text and accurately represent the content.
\end{enumerate}

Here's an example using a sample text about the benefits of a healthy diet:

Text: "A healthy diet consists of various food groups, such as fruits, vegetables, whole grains, lean proteins, and healthy fats. Consuming a balanced diet can help maintain a healthy weight, support the immune system, and reduce the risk of chronic diseases."

Conversation:

User: What are the main food groups in a healthy diet?

Assistant: A healthy diet includes various food groups, such as fruits, vegetables, whole grains, lean proteins, and healthy fats.

User: How can a healthy diet help maintain a healthy weight?

Assistant: A balanced diet provides the essential nutrients your body needs without excessive calories. Consuming the right balance of foods can help you achieve and maintain a healthy weight by preventing overeating and ensuring proper energy intake.

User: What are the benefits of a healthy diet for the immune system?

Assistant: A healthy diet supports the immune system by providing essential vitamins, minerals, and antioxidants that help the body fight off infections and maintain overall health.

User: How does a balanced diet reduce the risk of chronic diseases?

Assistant: Consuming a balanced diet can lower the risk of chronic diseases by providing the necessary nutrients for optimal body function and reducing the intake of harmful substances like saturated fats, trans fats, and added sugars. A healthy diet can help prevent conditions such as heart disease, diabetes, and certain types of cancer.

\end{oframed}

\begin{oframed}
\textbf{PROMPT3:}

As an AI assistant, your task is To generate question types, questions, answers, and source text for answers that comprehensively cover a given text and sample questions present them in a table format, follow these steps:

\begin{enumerate}
    \item Read the given text: Carefully read the text to understand its content and context.
    \item Identify key information: Look for definitions, metrics, facts, figures, events, or statements that can be used to create questions, ensuring that all parts of the text are covered.
    \item Generate questions: Create questions according to the chosen question types, ensuring they are clear, concise, and aligned with the question type. Make sure the questions cover the entire text.
    \item Locate answers: Find the corresponding answers for the questions within the text.
    \item Obtain source text: Replicate the essential text sections that offer the solutions to the queries, functioning as the basis for the responses. The source text must be thoroughly exhibited in a complete manner. It is necessary for the source text to be presented in a clear and understandable English without utilizing ellipsis.
    \item Design a table: Create a table with columns for question type, question, answer, and source text.
    \item Populate the table: Fill the table with the generated question types, questions, answers, and the extracted source text.
    \item Review and revise: Double-check the table for accuracy, comprehensiveness, and make any necessary adjustments.
\end{enumerate}

Here's Sample Questions:

\begin{itemize}
    \item Can you describe the responsibilities of the <committee name> and the <committee name> under the <entity name>?
    \item What are the categories of <entity group name>, and could you provide examples of entities falling under these categories?
    \item Who are the stakeholders that must abide by the principles outlined in the Governance Manual, according to the text?
    \item Can you provide a comprehensive list of all the authorized documents that are recognized by the Governance Manual and contribute to its overall objectives and goals?
    \item Where is <entity name> in the Governance Framework?
    \item What is <entity name>?
    \item What are the entities within the <category name> of the Framework?
\end{itemize}

Here's an example of how to structure your table:

| Question Type | Question | Answer | Source Text |

|------------------------|----------|--------|----------------------|

| Dialog | Q1 | A1 | Source Text 1 |

| Conversation | Q2 | A2 | Source Text 2 |

| Conversation Summary | Q3 | A3 | Source Text 3 |

| Conversation Fact Check| Q4 | A4 | Source Text 4 |

| Fact Check | Q5 | A5 | Source Text 5 |

| True or False | Q6 | A6 | Source Text 6 |

| Summary | Q7 | A7 | Source Text 7 |

| Quiz | Q8 | A8 | Source Text 8 |

| Review | Q9 | A9 | Source Text 9 |

| Fill in the Blank | Q10 | A10 | Source Text 10 |

| Short Answer | Q11 | A11 | Source Text 11 |

| Yes or No | Q12 | A12 | Source Text 12 |

| Matrix | Q13 | A13 | Source Text 13 |

| Compound Question | Q14 | A14 | Source Text 14 |

| ... | ... | ... | ... |

Remember to generate a variety of question types, questions, answers, and source text, and aim for a total of 10 questions or more questions.

\end{oframed}

\subsection{System Prompt for inference}
\label{appx:system_prompt}

\begin{oframed}

As a <organization document name> assistant, Your job is provide with accurate information  exclusively based on the <organization document name> and utilizing the given context. Only use the context if it is relevant to the question.

Follow the below instructions to answer questions in a step-by-step manner:

\begin{enumerate}
    \item Carefully analyze the question to comprehend what is being asked. Pay attention to the keywords, indirect cues, intentions, and specific details mentioned in the question.
    \item Find relevant information within the <organization document name> and provided context to construct your answer. If the given context is not suitable, do not use it.
    \item Support your answer with facts or information from the <organization document name> and provided context. If the given context is not relevant, do not need to use it.
    \item Organize the main points in a concise and clear manner.
    \item DO NOT create answers that aren't in the <organization document name>.
    \item Provide brief and clear response to the question, focusing on main point ONLY.
    \item If you can't answer a question, politely state that you're unable to do so.
\end{enumerate}

Please take note that <organization acronym> and <organization name> have identical meanings. Additionally, phrases such as "types", "departments", "entities", "categories", "positions", "groups", "units", "classifications" and "directorates" might also suggest the same idea. Furthermore, any significant keywords like "All" or "comprehensive" imply that the information data is sourced only from the manual.

Remember, as an AI assistant, it is vital to provide precise and dependable information based on the <organization document name>.
        
\end{oframed}

\section{Code Snippets for Entity Tree}

This sections shows the code for defining the entity parser component in the spaCy library and the code for parsing entities from user queries and performing the tree search.

\begin{oframed}

\textbf{Code Snippet 1:}

Defining a rule-based entity name matching rule in spaCy.

\begin{lstlisting}
def enhance_spaCY(ent_name):
    nlp = spacy.load("en_core_web_sm")
    ruler = nlp.add_pipe("span_ruler")

    patterns=[]

    for t in ent_name:
        temp = t.split()
        if len(temp) > 1 and t.find(',')<0 and t.find('&')<0:
            
            temp_pattern = []
            for a_temp in temp:
                temp_pattern.append({"LOWER": a_temp.lower()})
                        
            key_id = find_key_entity(t)
            
            patterns.append({"label": "UNHCRORG", "pattern": temp_pattern, 
                             "id": key_id.lower()})
        else:
            key_id = find_key_entity(t)
            
            patterns.append(
                {"label": "UNHCRORG", "pattern": t.lower(), 
                "id": key_id.lower()})

    ruler.add_patterns(patterns)

    return nlp


\end{lstlisting}

\end{oframed}

\begin{oframed}
\textbf{Code Snippet 2:}

Entity parsing from the user's query. spaCy library extracts named entities based on the rule defined earlier. The tree context can then be generated for the entities mentioned in the user's query and returned as context information.

\begin{lstlisting}
def search_entity_info(tree, nlp, search):
    search_context=[]
    search = search.lower().strip()

    doc = nlp(search)
    for span in doc.spans["ruler"]:
        if span.label_ == 'UNHCRORG':
            search_context.append(get_node_info(tree, span.id_))
    
    return '\n'.join(search_context)

\end{lstlisting}
\end{oframed}

\end{document}